\documentclass[journal]{IEEEtran}

\usepackage{cite}

\usepackage{graphicx,type1cm,eso-pic,color}

\usepackage{psfrag}

\usepackage{subfigure}

\usepackage{palatino, url, multicol}

\usepackage[lined,algonl,boxed]{algorithm2e}

\usepackage{setspace}

\begin{document}


\title{Controlling Wheelchairs by Body Motions: A Learning Framework for the Adaptive Remapping of Space}
\author{{Tauseef~Gulrez~\IEEEmembership{~Member~IEEE}, Alessandro~Tognetti, Alon~Fishbach, Santiago~Acosta,
Christopher~Scharver, Danilo~De~Rossi and Ferdinando~A.~Mussa-Ivaldi~\IEEEmembership{Member~IEEE,}}
\thanks{T.~Gulrez is with Robotics Lab. of Sensory Motor Performance Program, Rehabilitation Institute 
of Chicago, Feinberg School of Medicine, Northwestern University, Chicago, USA \& also with the Virtual and Interactive Simulations of Reality (VISOR) Labs, Department of Computing, Division of Information and Communication Sciences, 
Macquarie University Sydney,Australia}
\thanks{A.~Tognetti and D.De~Rossi are with Inter-Departmental Research 
Center
``E.Piaggio", University of Pisa, Italy.}
\thanks{F.A.~Mussa-Ivaldi, A.~Fishbach, S.~Acosta and C.~Scharver are with
Robotics Lab. of Sensory Motor Performance Program, Rehabilitation Institute 
of Chicago, Feinberg School of Medicine, Northwestern University, Chicago, 
USA.}}

\makeatletter
          \AddToShipoutPicture{
            \setlength{\@tempdimb}{.5\paperwidth}
            \setlength{\@tempdimc}{.5\paperheight}
            \setlength{\unitlength}{1pt}
            \put(\strip@pt\@tempdimb,\strip@pt\@tempdimc){
        \makebox(0,0){\rotatebox{55}{\textcolor[gray]{0.85}
        {\fontsize{3.0cm}{3.0cm}\selectfont{In Proc. Cogsys 2008}}}}
            }
        }
\makeatother

\maketitle

\begin{abstract}
Learning to operate a vehicle is generally accomplished by forming a new cognitive map between the body motions and extrapersonal space. Here, we consider the challenge of remapping movement-to-space representations in survivors of spinal cord injury, for the control of powered wheelchairs. Our goal is to facilitate this remapping by developing interfaces between residual body motions and navigational commands that exploit the degrees of freedom that disabled individuals are most capable to coordinate. We present a new framework for allowing spinal cord injured persons to control powered wheelchairs through signals derived from their residual mobility. The main novelty of this approach lies in substituting the more common joystick controllers of powered wheelchairs with a sensor shirt. This allows the whole upper body of the user to operate as an adaptive joystick. Considerations about learning and risks have lead us to develop a safe testing environment in 3D Virtual Reality. A Personal Augmented Reality Immersive System (PARIS) allows us to analyse learning skills and provide users with an adequate training to control a simulated wheelchair through the signals generated by body motions in a safe environment. We provide a description of the basic theory, of the development phases and of the operation of the complete system. We also present preliminary results illustrating the processing of the data and supporting of the feasibility of this approach. 
\end{abstract}
\IEEEpeerreviewmaketitle
\section{Introduction}
\begin{figure*}
\centering
\includegraphics[width=1.5 \columnwidth]{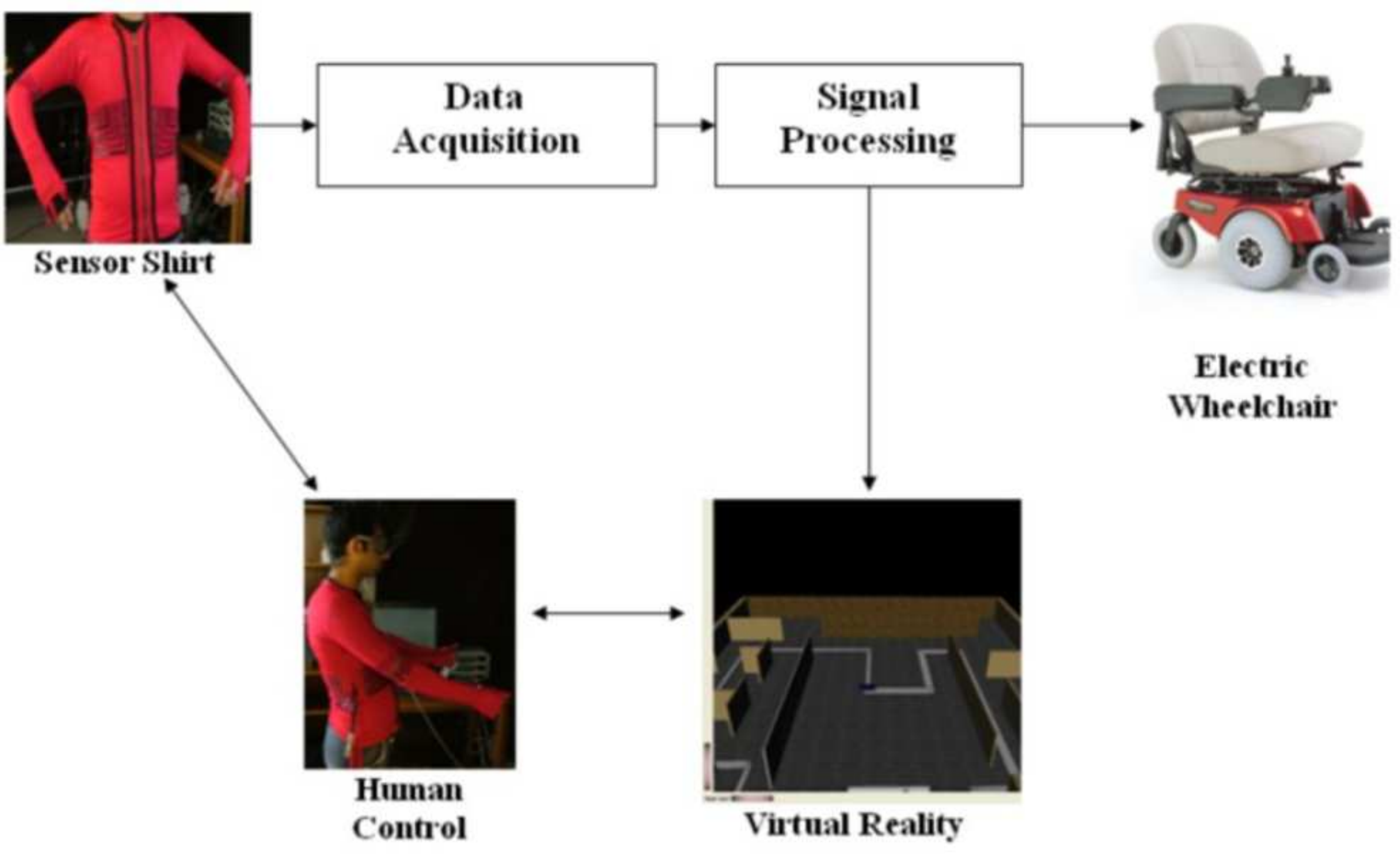}
\label{fig_02}
\caption{System concept. The virtual environment provides a safe training platform, where the control parameters are set according to the motor skills of the users. Once a satifactory behavior is reached, the control parameters will be applied to an actual powered wheelchair.}
\end{figure*}
\PARstart{R}{obotics} may be exploited to assist people in a great variety of activities~\cite{sandro_25,sandro_22,plats,sandro_01}. Elderly and disabled
people, in particular, are likely to benefit from these new technologies~\cite{plats,post,assist}. As they become limited in their mobility, they gain a greater degree of independence through the use of assistive devices such as powered wheelchairs(Fig.~\ref{fig_02}). However, loss of coordination and cognitive impairments can render difficult or impossible to execute steering maneuvers, with consequent fatigue, frustration, reduced social life and risks of dangerous accidents. One way to overcome these difficulties is to equip the chair with an intelligent controller, sharing planning and execution of actions with the user. This cooperation between human and machine can be compared to the cooperation between a horse and its rider: the rider navigates (global planning, ride control), while the horse avoids obstacles and makes path adjustments(fine motion control). A different approach - pursued here - is to allow the users to control the vehicle's motions at all levels. This second approach requires establishing a rapid communication between human and machine. However, one of the most challenging tasks does not concern the technology of communications and control, but rather the reorganization of movements and the development of new cognitive maps of motor space. This is something most of us are familiar with, as we learn to drive a car. At first, the controls are foreign objects that require constant focus and attention. But, as we become expert drivers, the car becomes an extension of our bodies and the acts that we perform on the steering wheel and the pedals are directly and seemlsessly mapped into into their spatial and temporal consequences. Here, we plan to achieve the same result in disabled populations through the interaction of human and machine learning. This paper desctibes the basic platform, which includes wearable sensing, interfacing and VR technologies.

  A typical powered wheelchair~\cite{cooper_01,cooper_02,robowheels_01,robowheels_02,robowheels_03,robowheels_04} is operated by two rear differential and two front castor wheels. Two high torque motors drive the rear wheels. Most powered wheelchairs come with a programmable joystick to drive and operate it. The joystick controller has four directional commands i.e. forward, backward, left and right and a zero position to halt the wheelchair operations. The velocity of the wheelchair incrementally increases up to a fixed limit by holding the joystick continuously in the desired direction. While the joystick is a simple control device, it still represents a fixed interface that the user must learn to operate by mapping joystick into wheelchair motions. Accidents are often caused by the insufficient training on the handling of the joystick and on the proper control procedures. Moreover, the wheelchair's training itself is a dangerous process, especially for spinal cord injured users.

 The need to apply learning technologies to control assistive devices is highlighted by a recent survey on the use of powered wheelchairs~\cite{fehr}. The authors interviewed 200 clinicians in spinal cord injury facilities, rehabilitation centers and geriatric care facilities. They asked about wheelchair user's feedback on the performance of different control interfaces, such as the joystick, sip-and-puff systems and head-an-chin devices. The study showed that about 10 percent of the disabled users ``find it extremely difficult or impossible'' to use the wheelchair while 40 percent of the users report difficulties in steering and maneuvering tasks. It is noteworthy that these figures refer to users that received specific (although conventional) training for controlling the wheelchair. In light of these difficulties, our approach is based on two characteristic features of the sensorimotor system:
\begin{itemize}
	\item Its ability to adapt to changes in the environment
and
\item Its ability to exploit a large number of degrees of freedom for carrying out a variety of tasks.
\end{itemize}
We exploit these features for designing a body-machine interface that will allow disabled users to operate a variety of devices~\cite{cochlea_06,sandro_01,sandro_02,sandro_26,cochlea_04,sandro_bmi}. In particular, we will aim at creating a learning and design framework for spinal cord injured people with complete injuries at the C5-6 cervical level, or incomplete injuries in the cervical cord. These injuries result in tetraplegia with limited residual body motions.

\section{Overall Methodology and System Description}
This article describes a novel method for controlling a powered wheelchair by spinal cord injured people. A wearable sensor shirt which is adequate to detect upper body (wrist, elbow and shoulder) movements, is custom built to extract some residual body movements of the users. A combination of virtual reality and signal processing methods is used for developing an effective body/device interface and for carrying out training procedures. The proposed system architecture is sketched in Fig. 1 and is based on four modules:

proposed system architecture is sketched in Fig.\ref{fig_02} and is based on three modules:
\begin{itemize}
	\item {\bf \em Sensor Shirt}: The sensor shirt is composed of 52 piezoresistive sensors that detect local fabric deformation caused by the movement of the user's upper body (i.e. wrist, elbow and shoulder; see section \ref{sec:shirt}).
	\item {\bf \em Data Acquisition and Signal Processing}: Signals acquired from the sensors are processed and the 
control parameters for the wheelchair control are determined (Section \ref{sec:sigacq}).
	\item {\bf \em Virtual Reality}: Preliminary system tests are performed on a virtual reality simulator of a powered wheelchair (section \ref{sec:vr}). The patient is trained to execute maneuvers of variable complexity, such as 
navigating a desert scene, moving among obstacles and following other moving objects.
  \item {\bf \em Human Control}Users are immersed in the PARIS system. After an initial calibration (Section V)
they begin practising the control of the simulated wheelchair. The signals generated by the shirt are transformed in command variables, which are integrated and combined with a head tracker (Flock of Birds, Ascension Technology~\cite{sandro_27,sandro_28,fobird_01}) to generate the current viewpoint from the simulated wheelchair.
\end{itemize}
\begin{figure}
\centering
\includegraphics[width=0.5 \columnwidth]{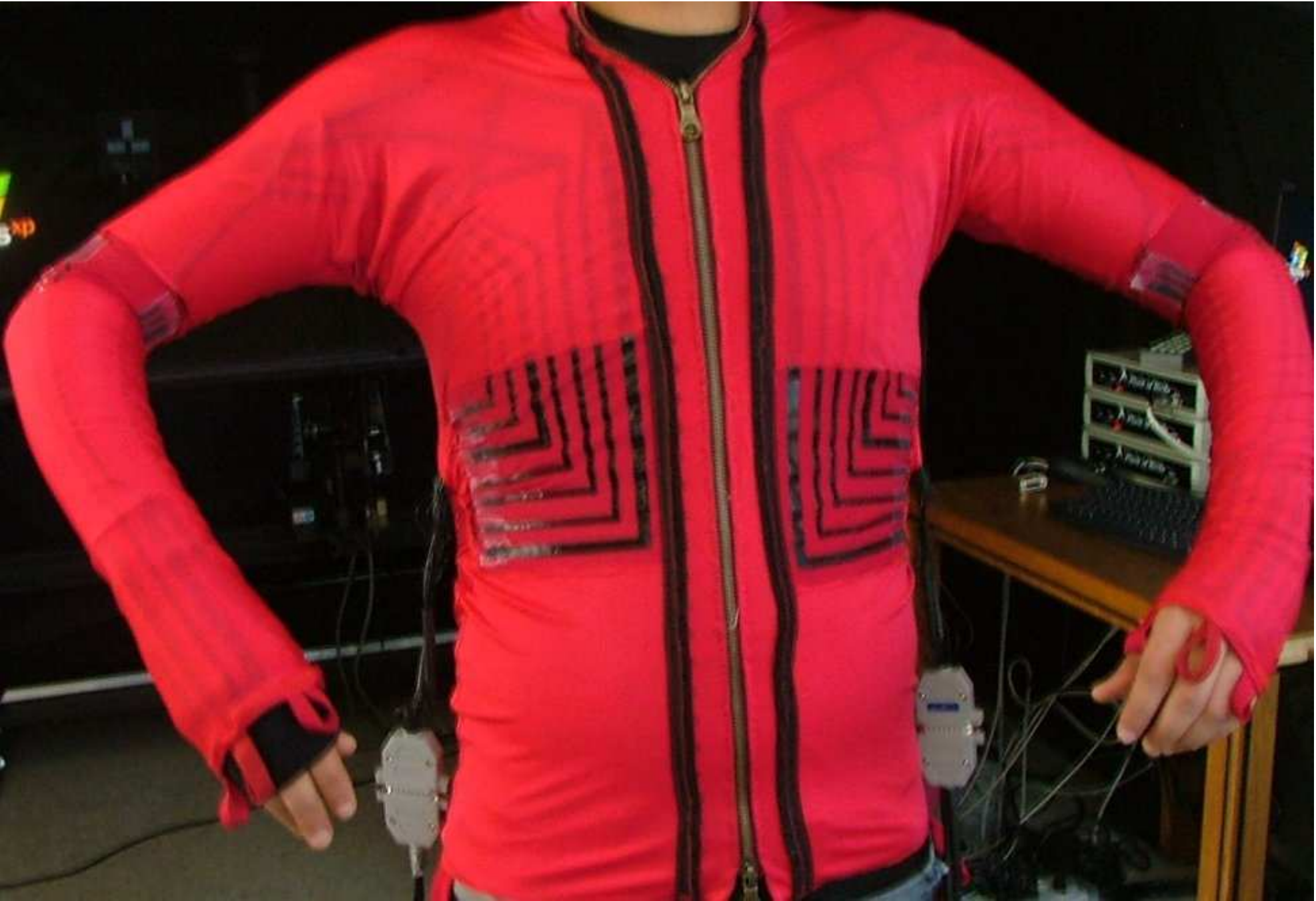}
\caption{Sensor Shirt front view}
\label{fig:shirt_1}
\end{figure}
\begin{figure}
\centering
\includegraphics[width=0.5 \columnwidth]{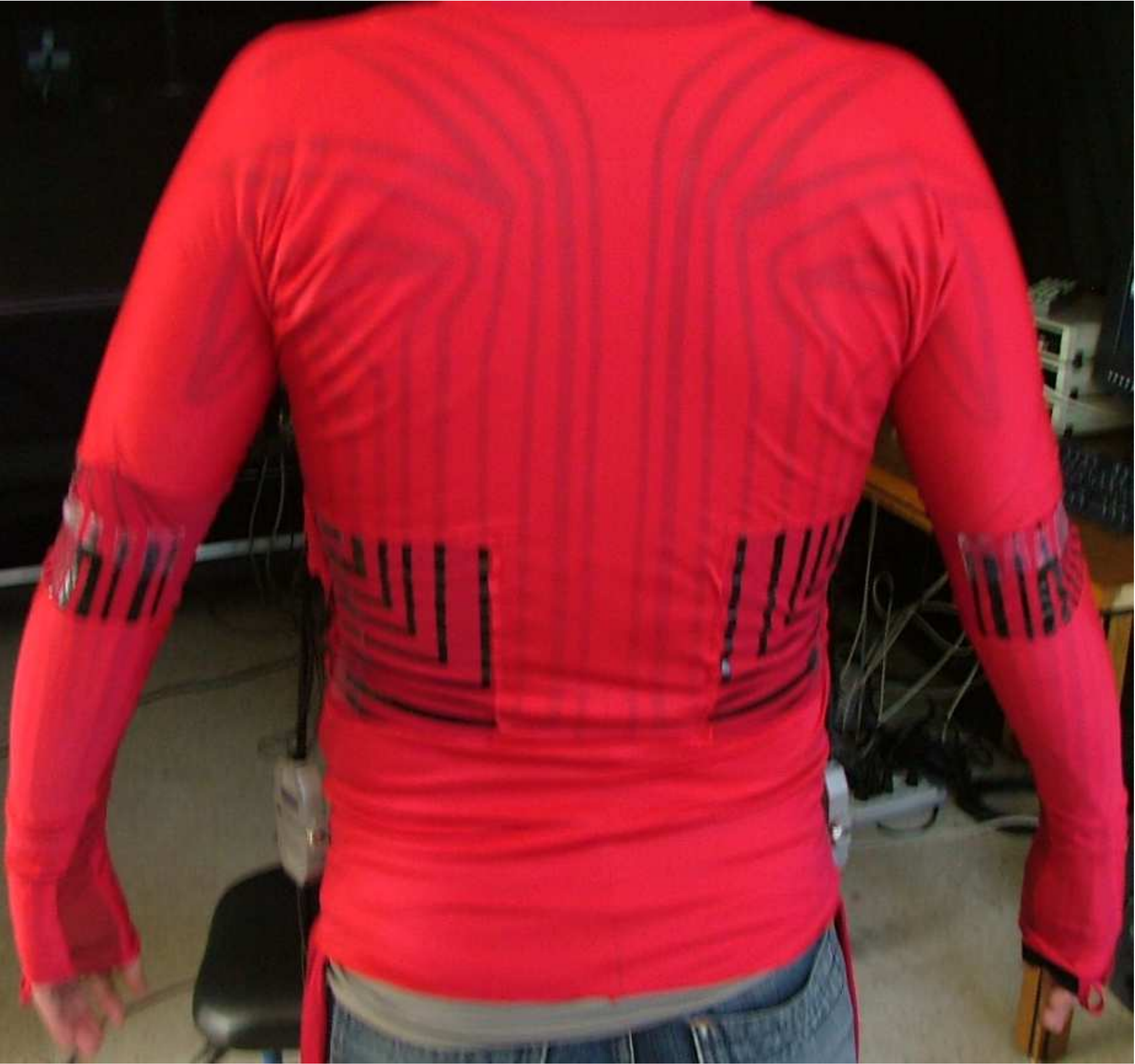}
\caption{Sensor Shirt back view}
\label{fig:shirt_2}
\end{figure}
\begin{figure}
\centering
\includegraphics[width=0.9\columnwidth]{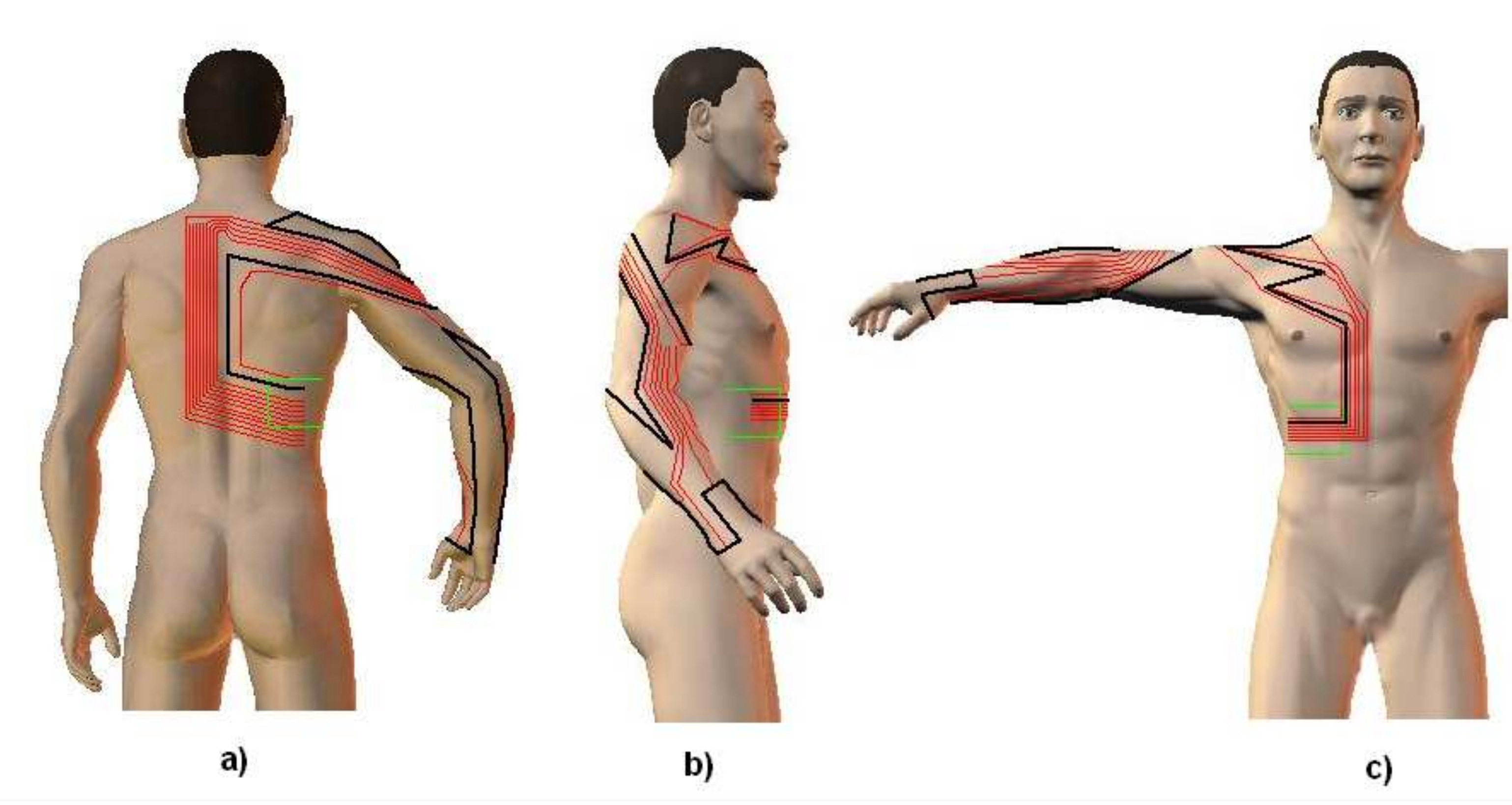}
\caption{a) Sensors on the back portion of the shoulder. b)Sensors on the muscle arm joint, elbow and wrist. c)Front view of the sesnors covering front shoulders and limb area. }
\label{sensor shirt layout}
\end{figure}
\section{Sensor Shirt} \label{sec:shirt}
The sensors of the shirt (Figures \ref{fig:shirt_1}, \ref{fig:shirt_2} and  \ref{sensor shirt layout}) are made of a conductive elastomer (CE) material (commercial product provided by Wacker LTD \cite{wak}) printed on a Lycra/cotton fabric previously covered by an adhesive mask. 

  CE composites show piezoresistive properties when a deformation is applied \cite{ piezo_poly2}. CE materials can be applied to fabric or to other flexible substrate, they can be employed as strain sensors \cite{lor2,hand} and they represent an excellent trade-off between transduction properties and possibility of integration in textiles. Quasi-statical and dynamical sensor characterization has been done in \cite{lor2}. CE sensors exhibit some non-linear dynamical properties and relatively long relaxation times \cite{piezo_poly,piezo_time} which should be taken into account in the control formulation.  
\subsection{Signal Acquisition} \label{sec:sigacq}
The analog signals acquired from the sensors are amplified and then digitized using a general purpose 64 channels acquisition card and real-time processed using a personal computer. Real-time signal processing has been performed by using the xPC-Target\textregistered toolbox of Matlab\textregistered. The output of the signal processing stage, i.e the wheelchair controls, are sent to the virtual wheelchair described in the section below by using UDP connection.
\section{Virtual Wheelchair and Personal Augmented Reality Immersive System
(PARIS)} \label{sec:vr}
\subsection{Software}
\begin{figure}
\centering
\subfigure[]
{\includegraphics[width=0.6 \columnwidth]{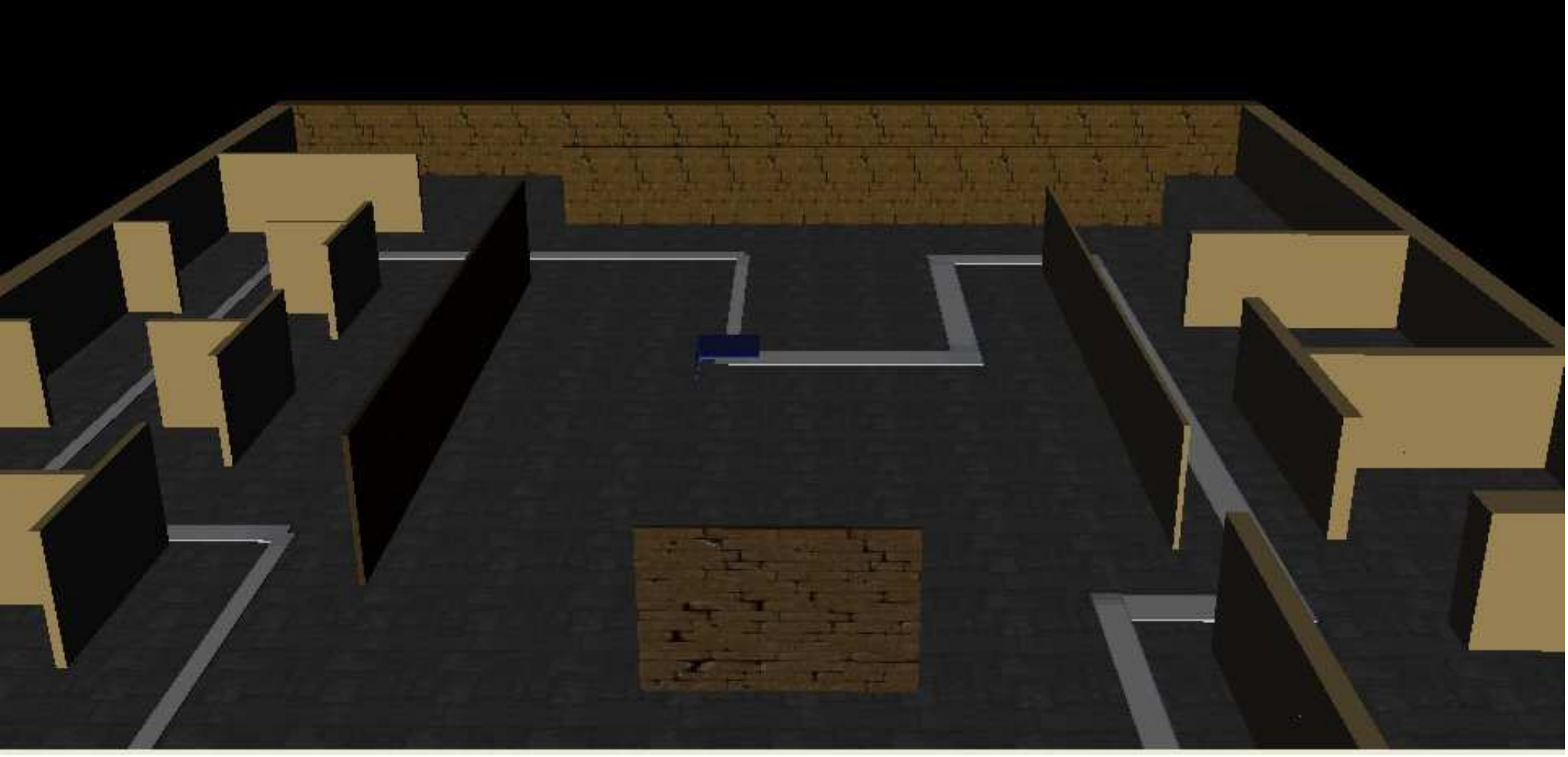}}
\subfigure[]
{\includegraphics[ width=0.45 \columnwidth]{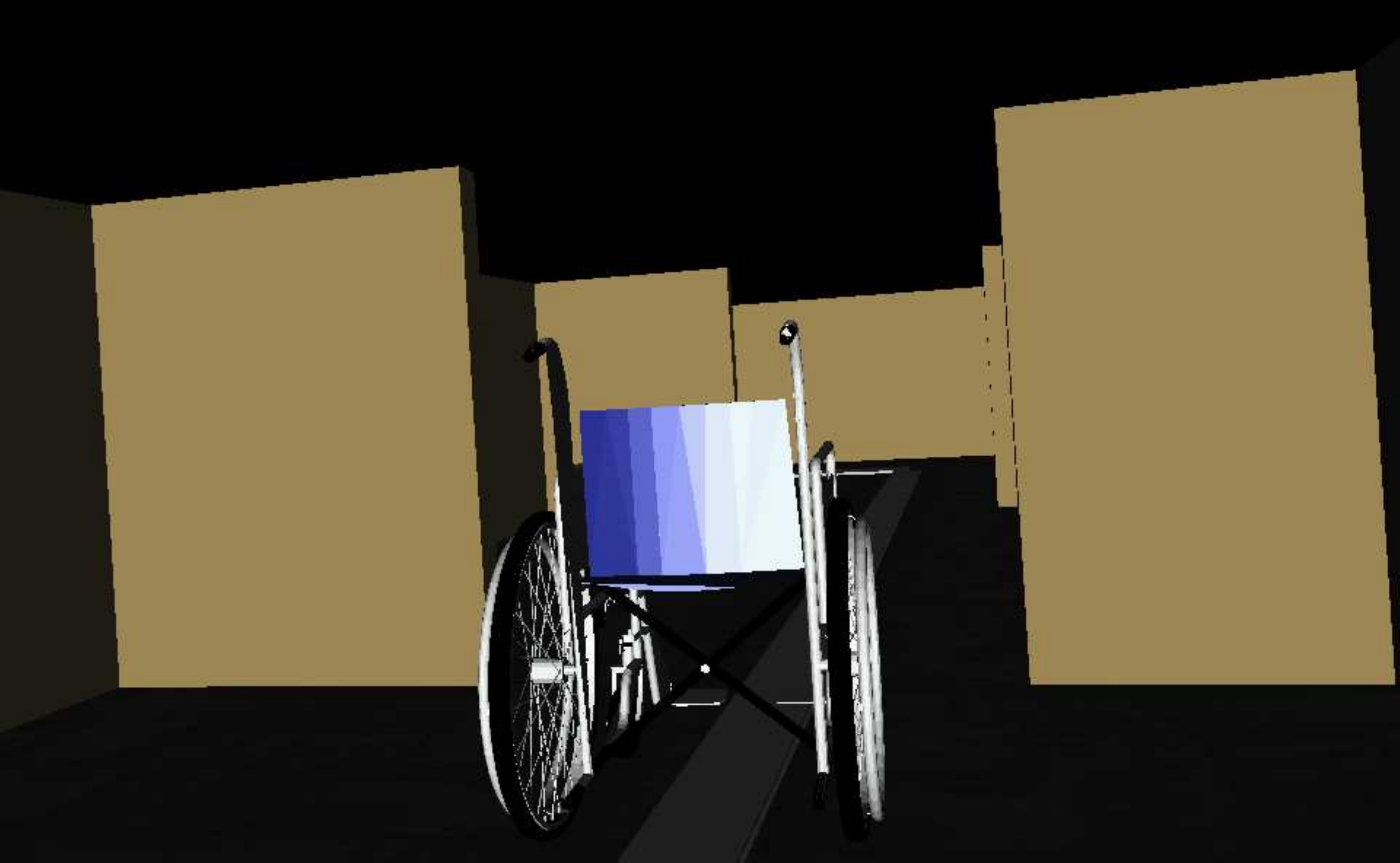}}
\subfigure[]
{\includegraphics[ width=0.45 \columnwidth]{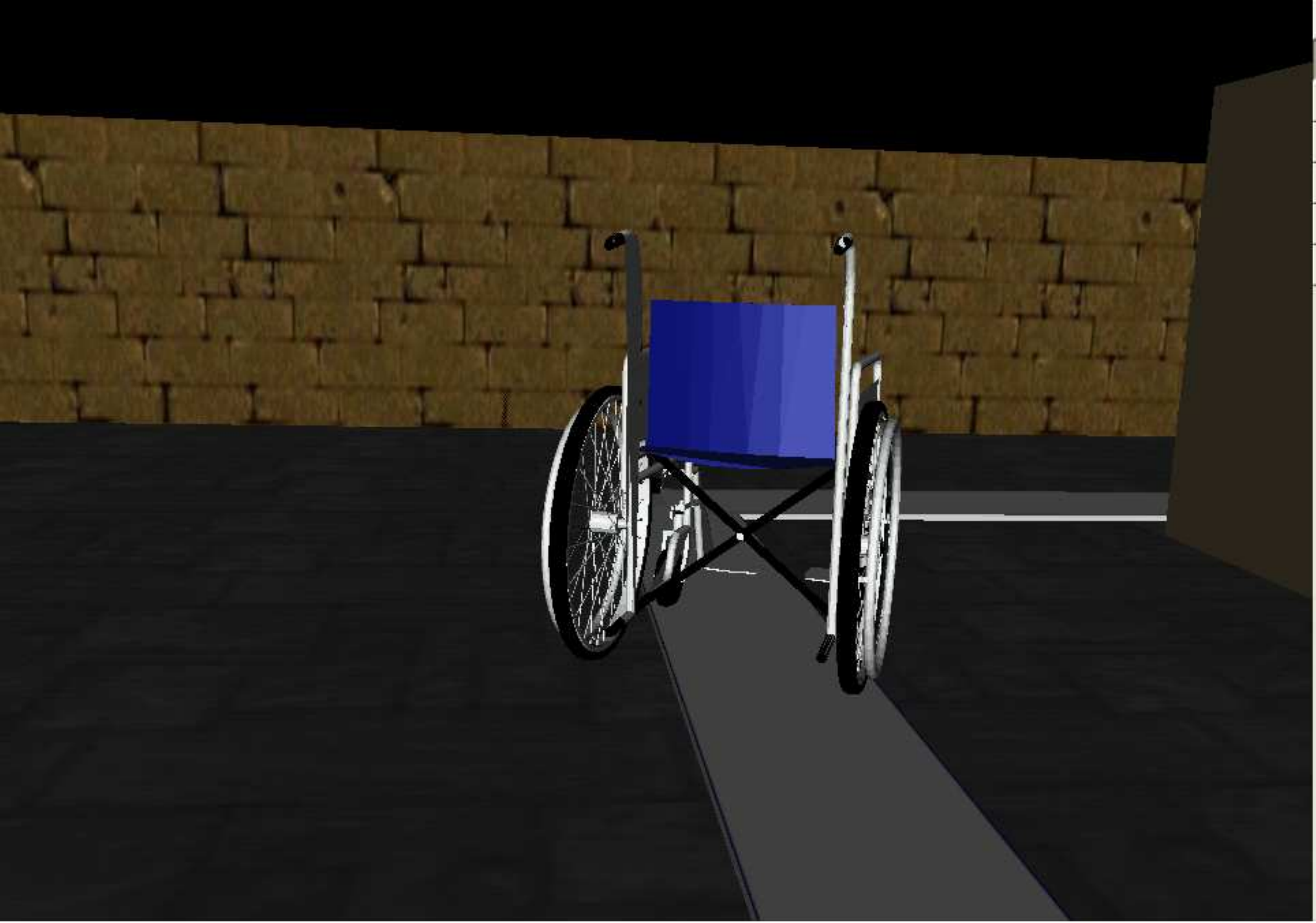}}
\caption{a) Virtual reality scene of our floor plan including corridors and 
small rooms which is projected in the PARIS. b)Robotics Virtual wheelchair 
is navigating through a door way. c)Patient is driving the Robotics 
wheelchair on
the marked path inside the virtual reality.}
\label{vr}
\end{figure}

The virtual wheelchair and its surrounding environment are designed using VRCO's CAVELib \texttrademark 3D Graphics \cite{vrco}, Coin3D graphics libraries and VRML models \cite{coin}. The whole program was simulated on a Personal Augmented Reality Immersive System (PARIS) as described in \cite{paris} and \cite{vrwheel}. PARIS provides the user with a perspective view of the scene. By wearing the specially designed goggles and head-tracker, users observe the
scene from the viewpoint of the moving wheelchair. The goggles are actively switched and synchronized with the projection system to provide 3D stereo vision of the artificially generated images. The scene is updated asynchronously, based on an external input from the sensor shirt and from a head mounted 3D tracker.
Fig.~\ref{vr} shows a layout for the virtual environment, composed of several corridors, walls, obstacles and doorways. A path (represented by the white track of Fig.\ref{vr}) is drawn on the floor as guide for the subject to track during the learning phase (Section \ref{sec:exp}).
\subsection {Virtual Wheelchair Kinematics Model}
\begin{figure}
\centering
\subfigure[]
{\includegraphics[width=0.43 \columnwidth]{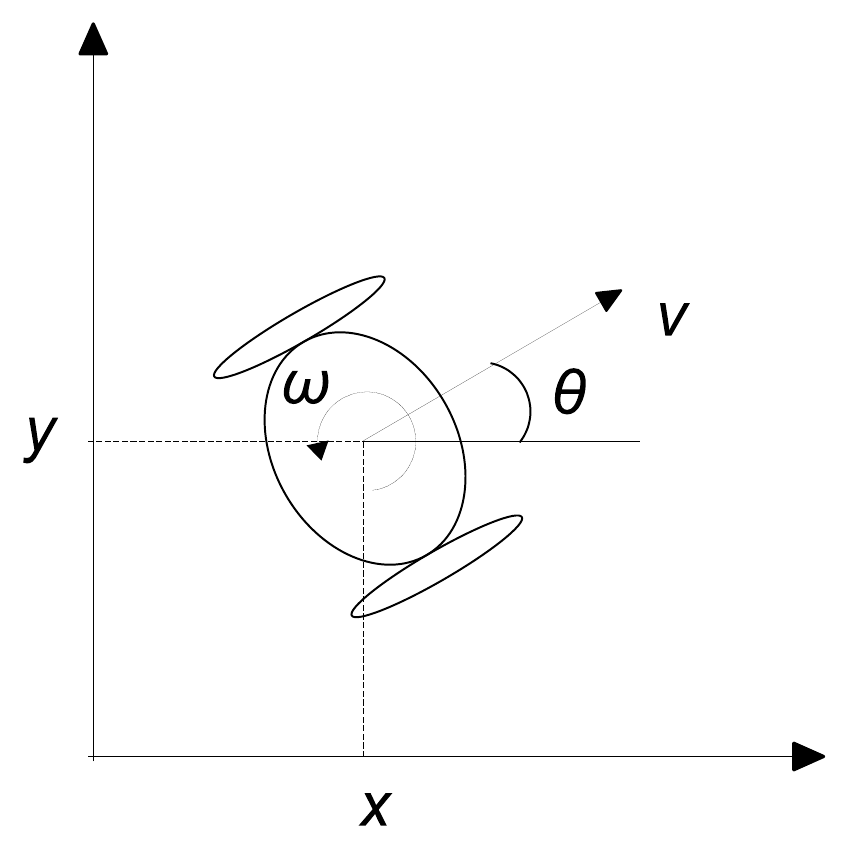}}\quad
\subfigure[]
{\includegraphics[width=0.43 \columnwidth]{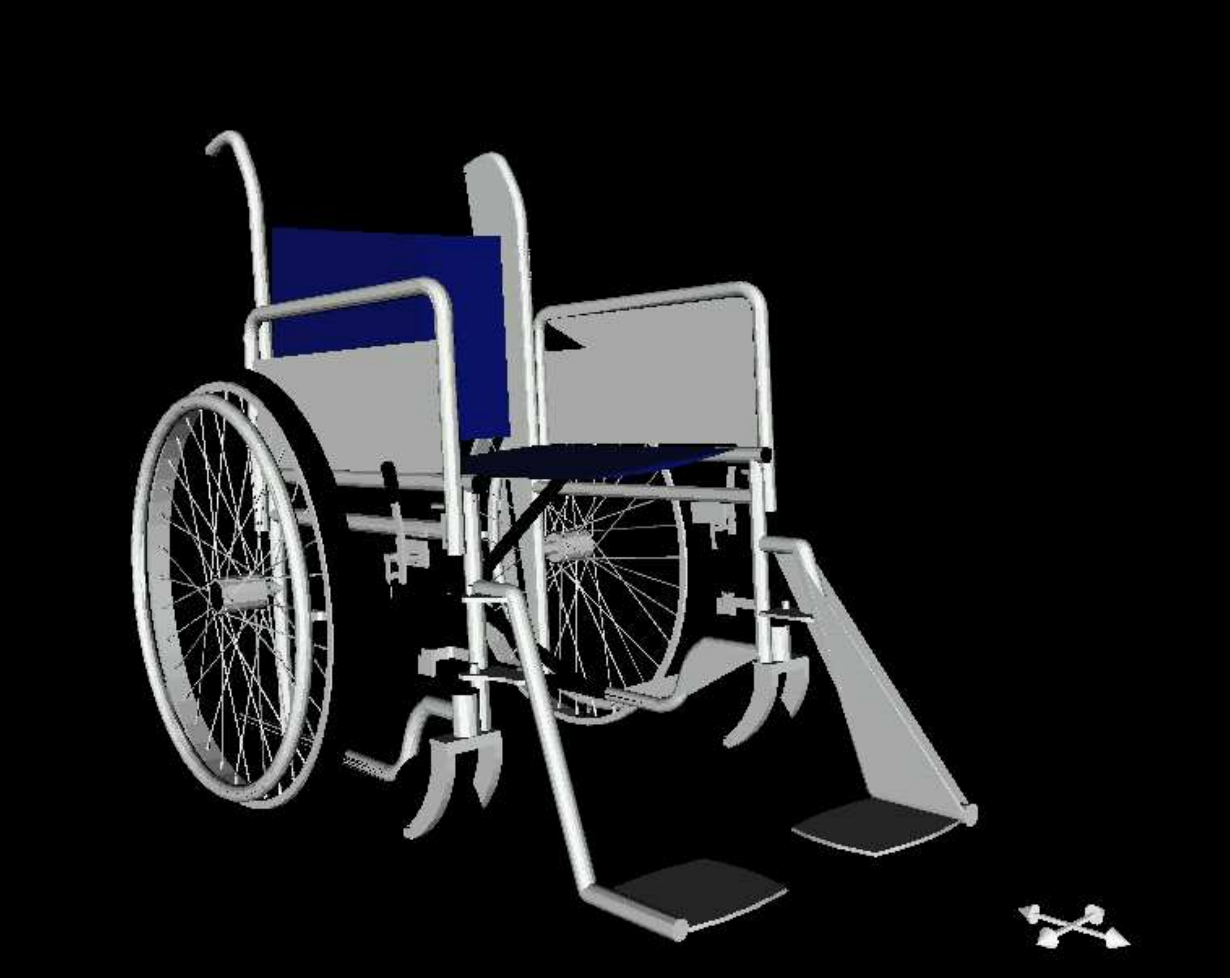}}\quad
\caption{a) Virtual wheelchair kinematics model based upon unicycle robot.
b)Virtual wheelchair's 3D model created using Coin3d\cite{coin} libraries.}
\label{kinmodel}
\end{figure}

\begin{figure}
\centering
\includegraphics[width=0.55 \columnwidth]{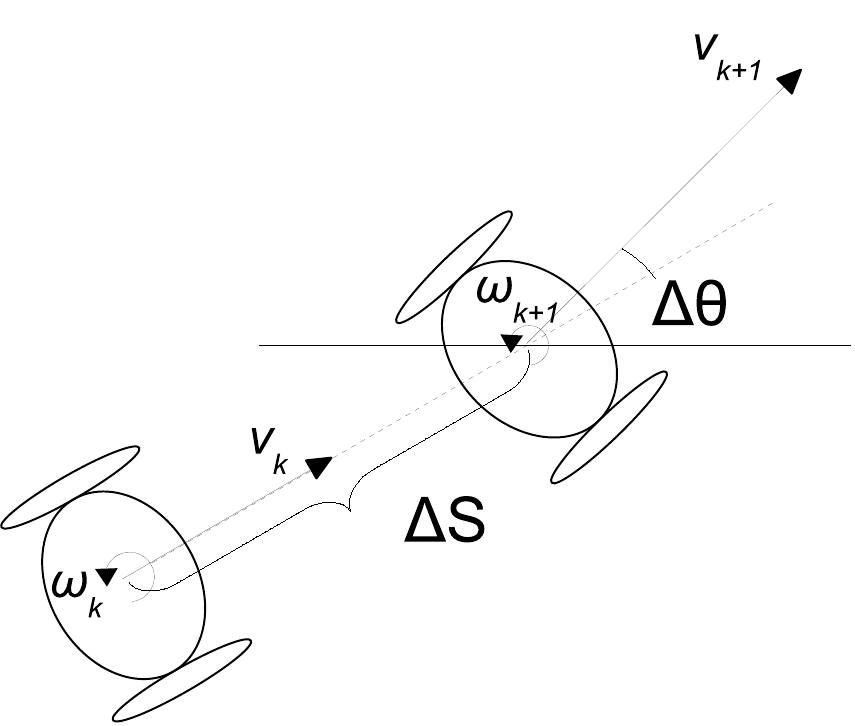}\quad
\caption{Virtual wheelchair's position update.}
\label{motion}
\end{figure}

The wheelchair is modeled as a simple two-wheel vehicle \cite{unicycle,satoshi}, as shown in Fig. \ref{kinmodel}. The non-holonomic kinematic equations of the wheelchair are:
\begin{eqnarray}
\nonumber \dot{x}(t) &=&\textit{v(t)}cos(\theta(t)) \label{kin}  \\
\dot{y}(t) &=&\textit{v(t)}sin(\theta(t))\\
\nonumber \dot{\theta}(t) &=& \omega(t)
\end{eqnarray}
The kinematic model of the wheelchair has two inputs, the translational 
velocity, {\em{v}} and the rotational velocity ($\omega$).
In discrete time, the wheelchair's laws of motion are:
\begin{eqnarray}
\nonumber x_{k+1}  =  x_{k}+\textit{v}_{k} cos(\theta_{k}) \Delta t  
\label{motionlow} \\
y_{k+1}  =  y_{k}+\textit{v}_{k} sin(\theta_{k} )\Delta t \\
\nonumber \theta_{k+1}= \theta_{k} + \omega_{k} \Delta t
\end{eqnarray}
The two control inputs, \textbf{$u_1$} and \textbf{$u_2$}, are generated by processing algorithms applied to the 
shirt signals (Section \ref{section:algo}). The virtual wheelchair position update from point $(x_{k},y_{k})$ to point $(x_{k+1},y_{k+1})$ (Fig. \ref{motion}) is given by:
\begin{eqnarray}
\Delta{S}  =  \textit{v}_{k} \Delta{t} = u_1 V_f \Delta{t} \\
\Delta{\theta} = \omega_{k} \Delta{t}  = u_2 V_r \Delta{t}
\end{eqnarray}
where $V_r$ and $V_f$ are the maximal rotational and forward velocities, respectively, and $\Delta{t}$ is the time 
interval between the two consecutive frames of the PARIS.
%
\section{Mapping body movements into Virtual Wheelchair (PARIS based) controls} \label{sec:mapbodymov}
\label{sec:mapp}
As a first stage in forming a map from body movements to wheelchair control, the nature of the control needs to be determined. The controls $\bf{u_1}$ and $\bf{u_2}$, for example, may specify the translational velocity {\em{v}} and rotational velocity ($\omega$). Alternatively, the controls may specify accelerations $\bf{\dot{v}}$ and $\bf{\dot{\omega}}$ instead of velocities. Given that we would like to allow the patient to remain in a fixed and comfortable position as much as possible we suggest to map one of the controls to the linear \textit{acceleration} of the wheelchair. This allows the patient to cruise at a fixed velocity while maintaining the resting posture.
\begin{figure}
\centering
\subfigure[]
{\includegraphics[width=0.99\columnwidth]{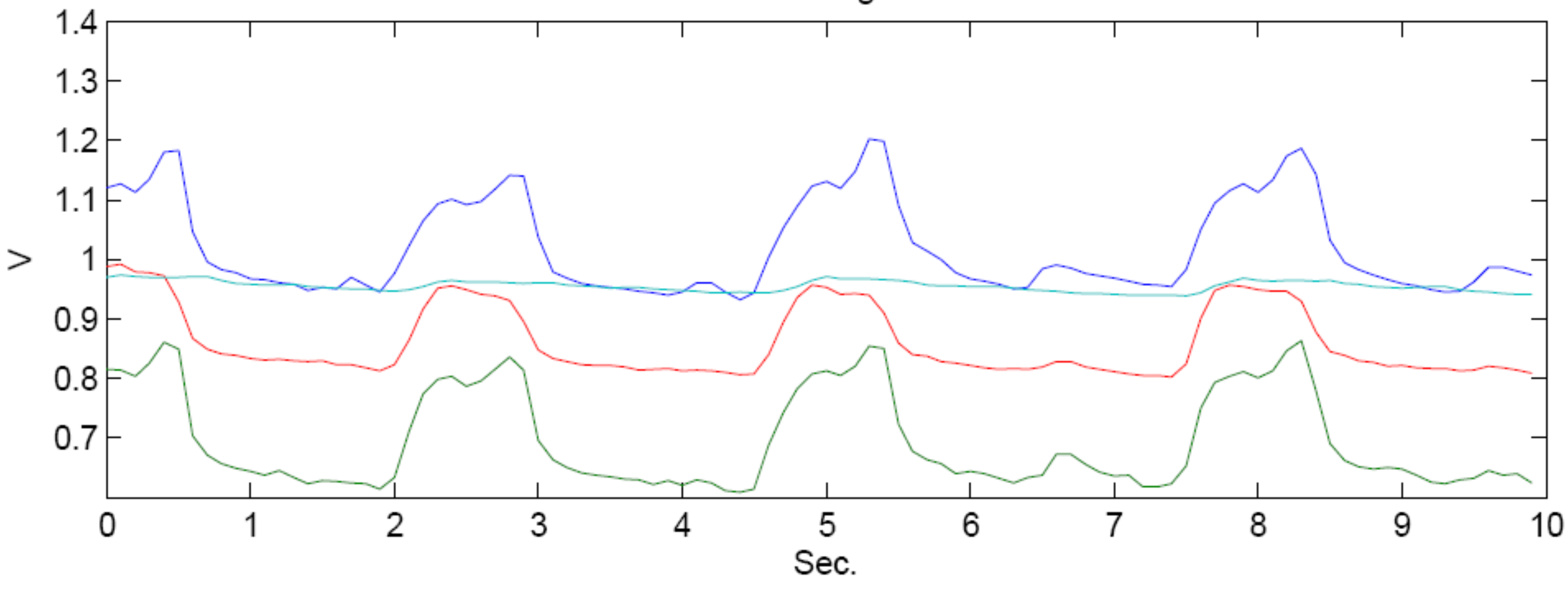}}
\subfigure[]
{\includegraphics[width=0.99\columnwidth]{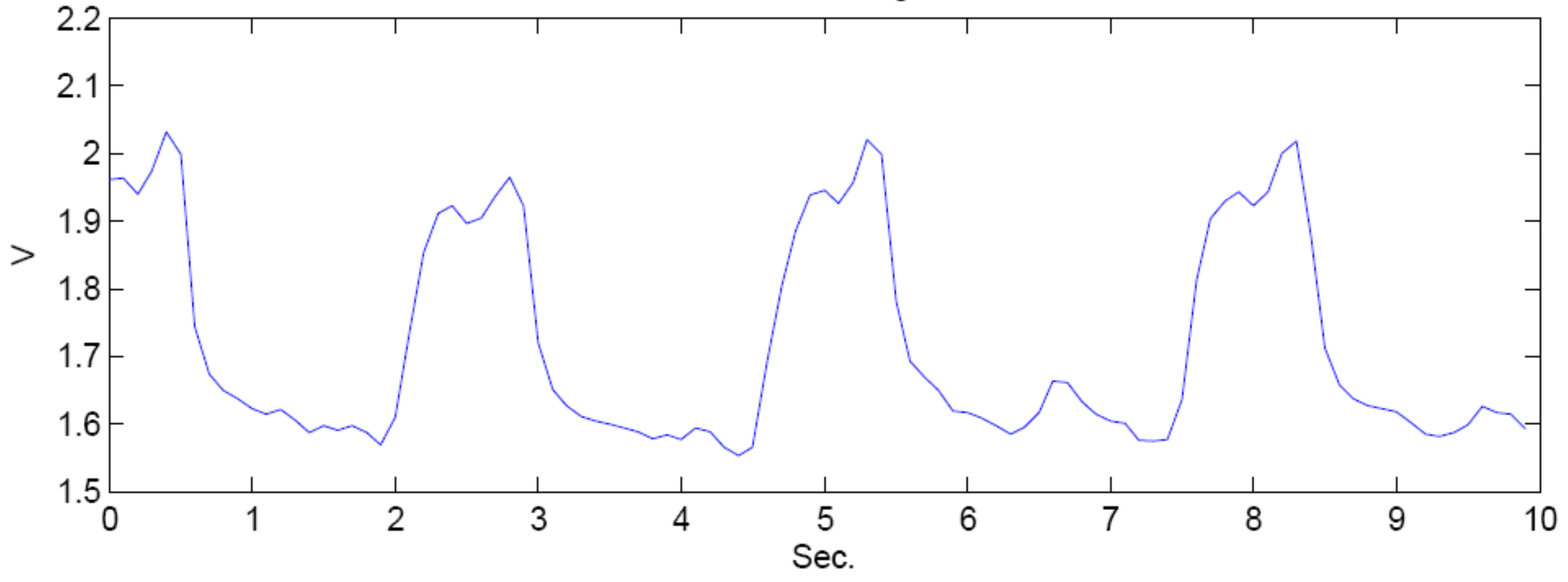}}
\caption{(a) Right elbow signals extracted during user elbow flexion. (b) First principal component extracted from the raw signals retaining the 80\% of the variance.}
\label{limbscalib}
\end{figure}
\begin{figure}
\centering
\includegraphics[height=120mm]{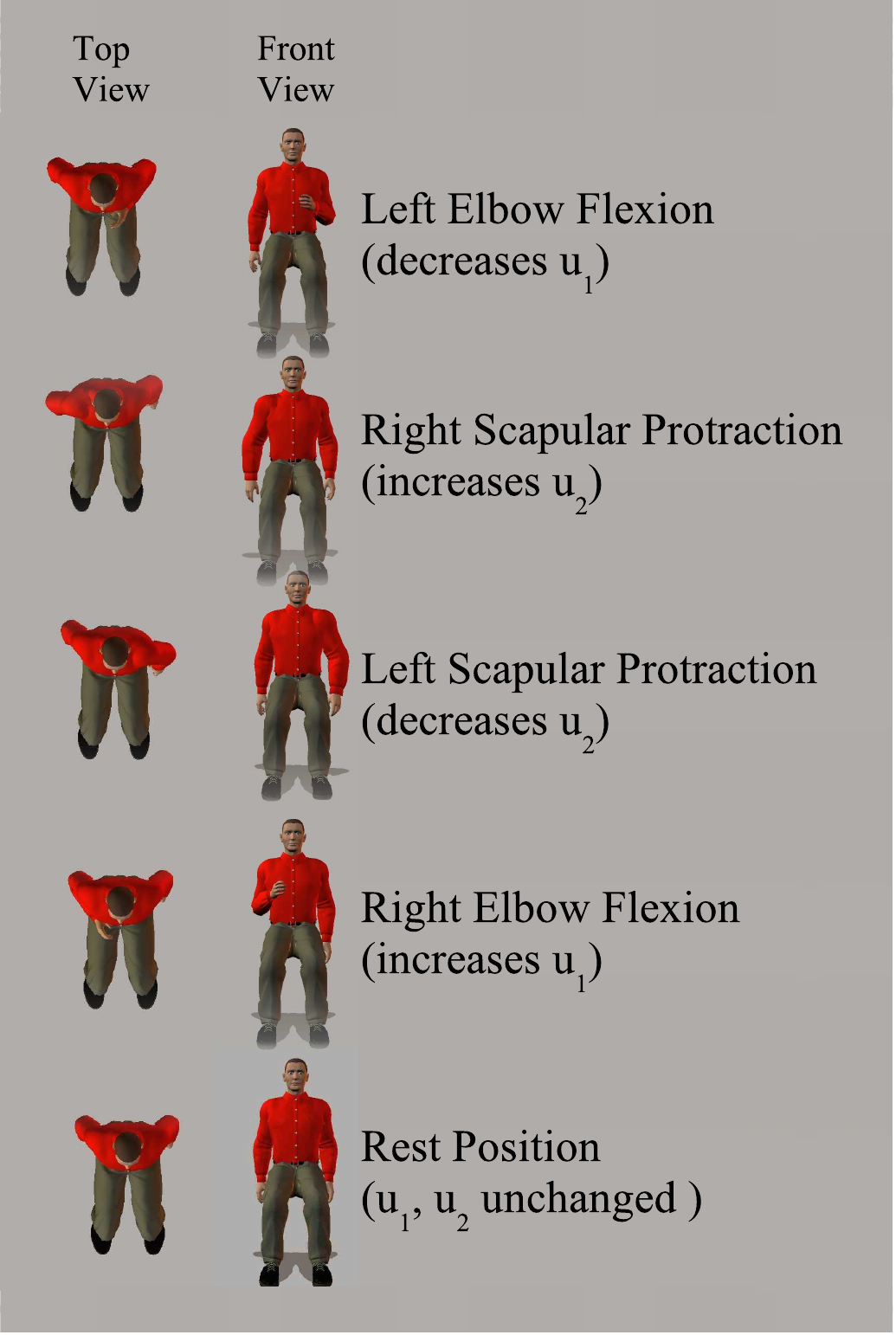}
\caption{Virtual Robotics Wheelchair's position update w.r.t. the body 
movements.}
\label{movement}
\end{figure}
Unlike the transitional velocity, the rotational velocity would typically be maintained at zero and would set to nonzero values for only short periods of time. For that reason we decided to map the second control signal to the rotational \textit{velocity}. In order to map the shirt signals to the controls one needs to assign certain body movements to each control. This allows for great flexibility, as the vocabulary is determined by the users, based on their specific movement ability and personal preferences. In our preliminary experiments, our subject decided to use the following body movements:
\begin{itemize}
	\item Right elbow flexion was used to increase the value of $\bf {u_1}$.
	\item Left elbow flexion was used to decrease the value of $\bf{u_1}$.
	\item Right shoulder movement forward (scapular protraction) was used to 
increase the value of $\bf{u_2}$.
	\item Left shoulder movement forward (scapular protraction) was used to 
decrease the value of $\bf{u_2}$. (see Fig. \ref{movement}).
\end{itemize}
Since the shirt contains several sensors at each joint, we examined the possibility of reducing the dimensionality
of the shirt signals by applying Principal Component Analysis (PCA) \cite{pca_01,pca_02,pca_07,pca_09} to the signals originating from the same joint. PCA was performed on data that were collected while the subject was moving his arms and
shoulders in an uninstructed manner for a period of 10 seconds. We found that the first principal component (PC) of each joint captures $80\%-90\%$ of the same-joint-sensors variance (see Fig.~\ref{limbscalib}). Thus, for the above control scheme, which was chosen by the subject, we use four signal combinations, the first PC of the right 
shoulder ($\bf{h_{rs}}$), the first PC of the left shoulder ($\bf{h_{ls}}$), 
the first PC of the right elbow ($\bf{h_{re}}$), and the first PC of the 
left elbow ($\bf{h_{le}}$).
\subsection{Description of Algorithms} \label{section:algo}
\begin{figure}
\centering
\subfigure[]
{\includegraphics[width=0.99 \columnwidth]{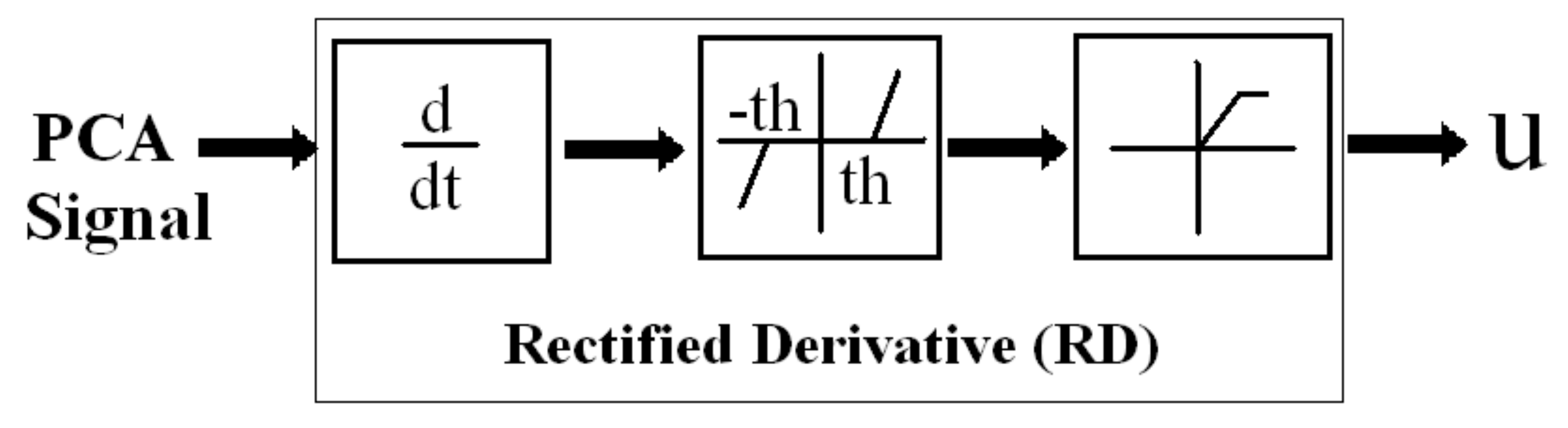}}
\subfigure[]
{\includegraphics[width=0.99 \columnwidth]{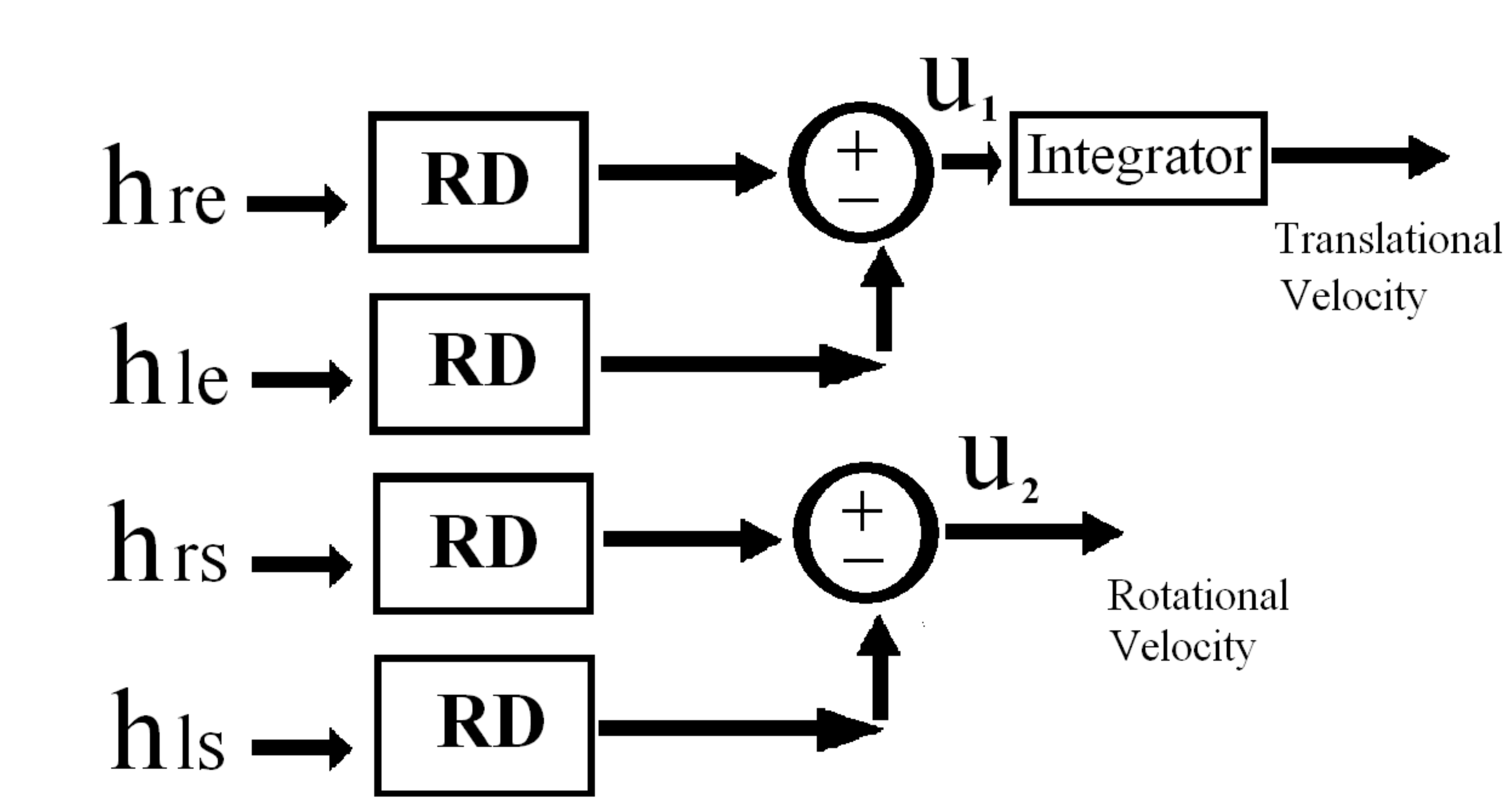}}
\caption{(a)Rectified Derivative Algorithm. (b)Control scheme block diagram.}
\label{signal}
\end{figure}

\begin{figure}
\centering
\includegraphics[width=0.99\columnwidth]{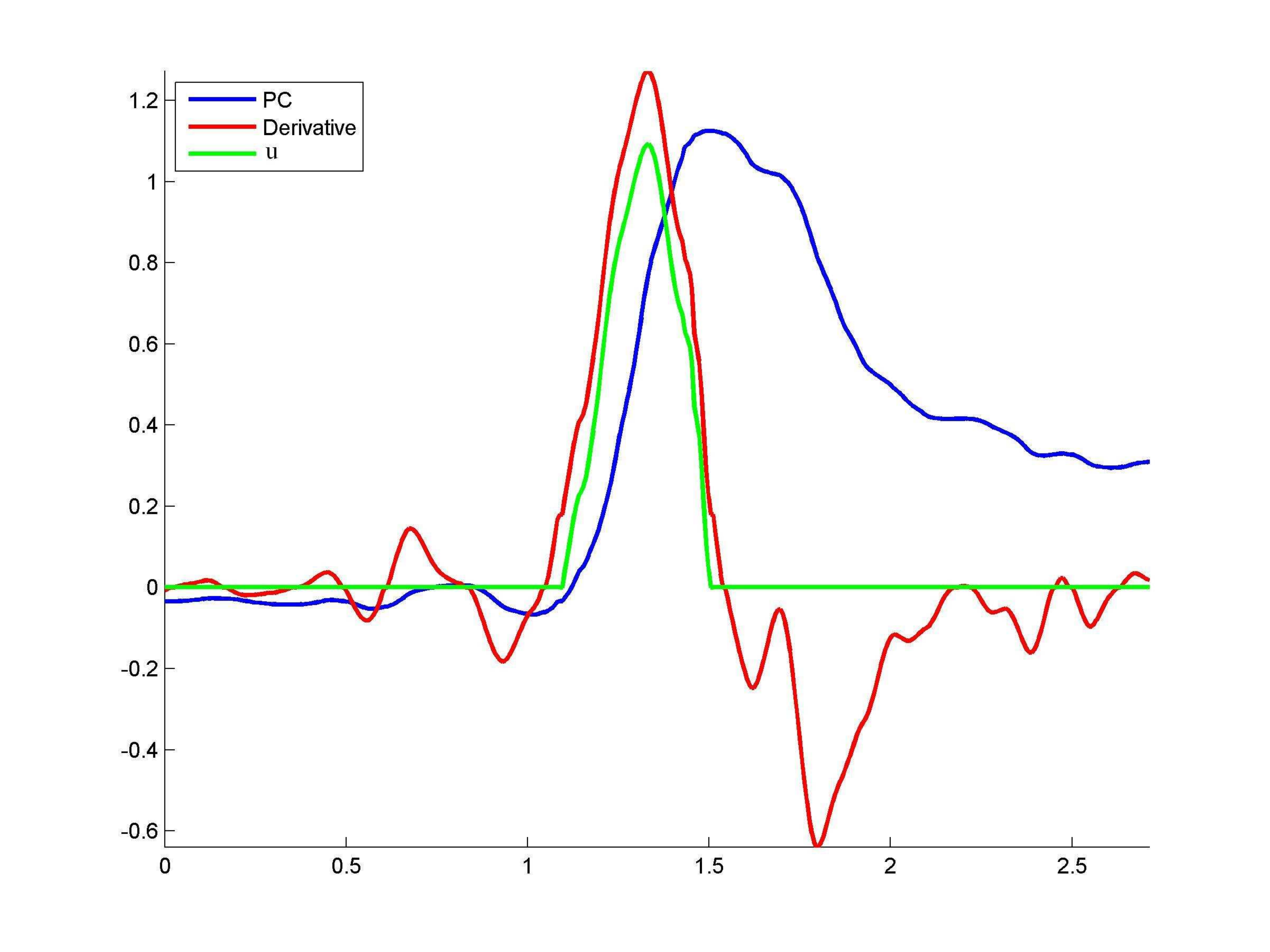}
\caption{Rectified derivative algorithm example.}
\label{proc2}
\end{figure}

For removing possible drift and noise artifacts from the shirt signals, we used the following algorithm. The
time derivative of each of the four PC's was calculated and a dead-zone was applied to each of them. The
signals were then positive-rectified, as we are only interested
in the rising part of each PC ((see Fig. \ref{signal}(a)). 
An example for the operation of the algorithm is shown in Fig. \ref{proc2}. 
The processed signals from the two elbows are then subtracted from each other to generate the transitional 
acceleration, while the processed signals from the two shoulders are subtracted from each other to generate the rotational velocity (see Fig. \ref{signal}(b)).
\section{Experimental Results} \label{sec:exp}
\subsection{Experimental Setup}
We present the results of a preliminary study conducted on a consenting adult participant approved by Northwestern University's Institutional Review Board (IRB). The participant wore a shirt, embedded with 52 piezo-resistive sensors, capable of detecting the wearer's residual mobility. With the shirt on, the subject was seated in front of a virtual reality system (discussed in section \ref{sec:shirt} \& \ref{sec:vr}). The virtual scene depicted in Fig.\ref{traj}(c) was modelled on a generic building floorplan, with multiple rooms, doors and corridors. A thick white line was marked on the floor and the subject was asked to navigate through the corridors and doorways following the white track Fig. \ref{traj}(d). The subject was able to navigate in the environment with little practice using arm and shoulder movements. Fig. \ref{traj}(a) shows the trajectory of the virtual wheelchair (red line) as the subject attempted to track the pathway (blue line). 
The raw shirt signals, extracted principal components and relative controls for the trajectory experiment are shown in Fig. \ref{segment}(c).
\subsection{Trajectory Analysis}
\begin{figure}
\centering
\subfigure[]{
\includegraphics[width=0.99\columnwidth]{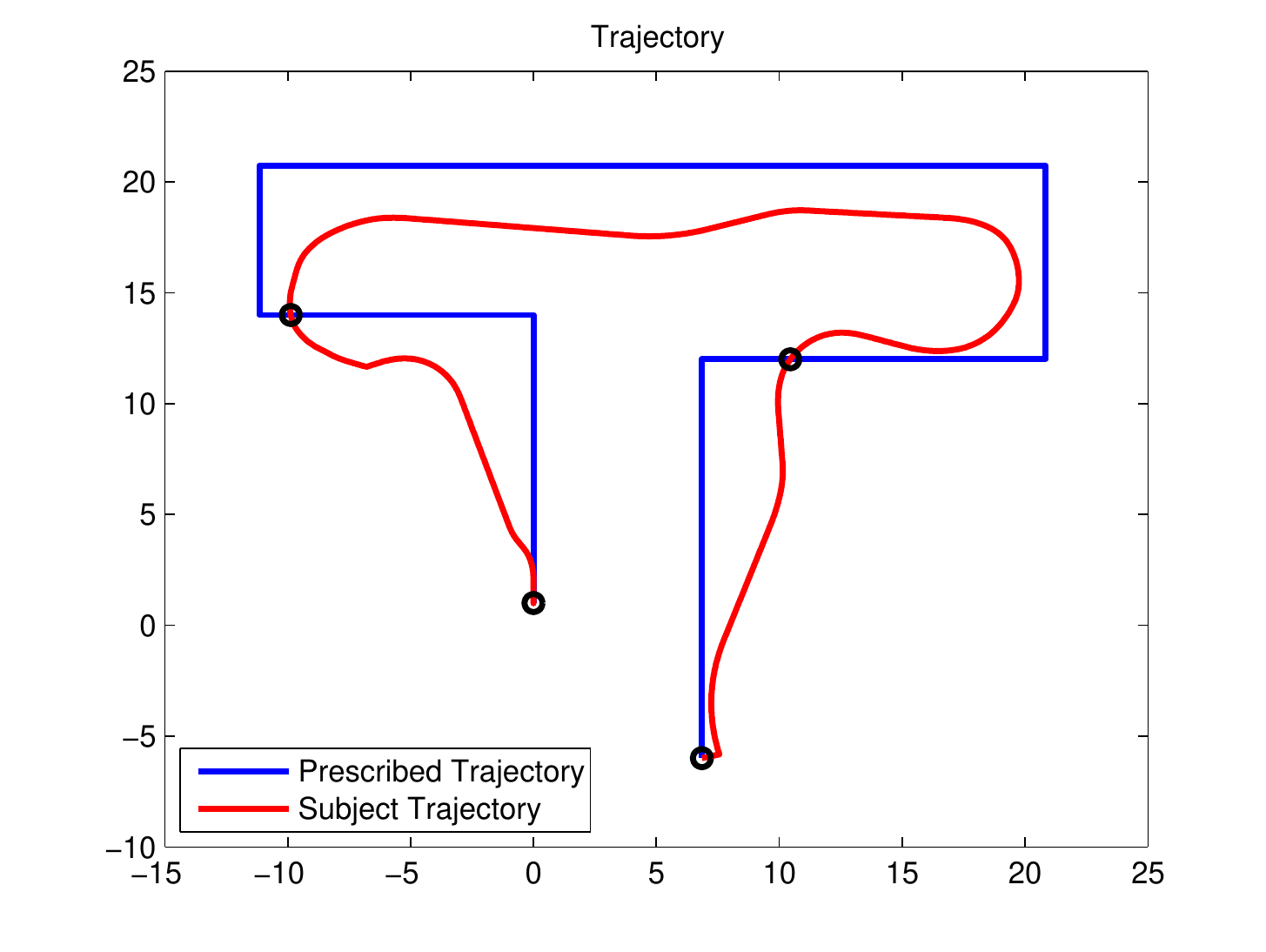}}
\subfigure[]{
\includegraphics[width=0.99\columnwidth]{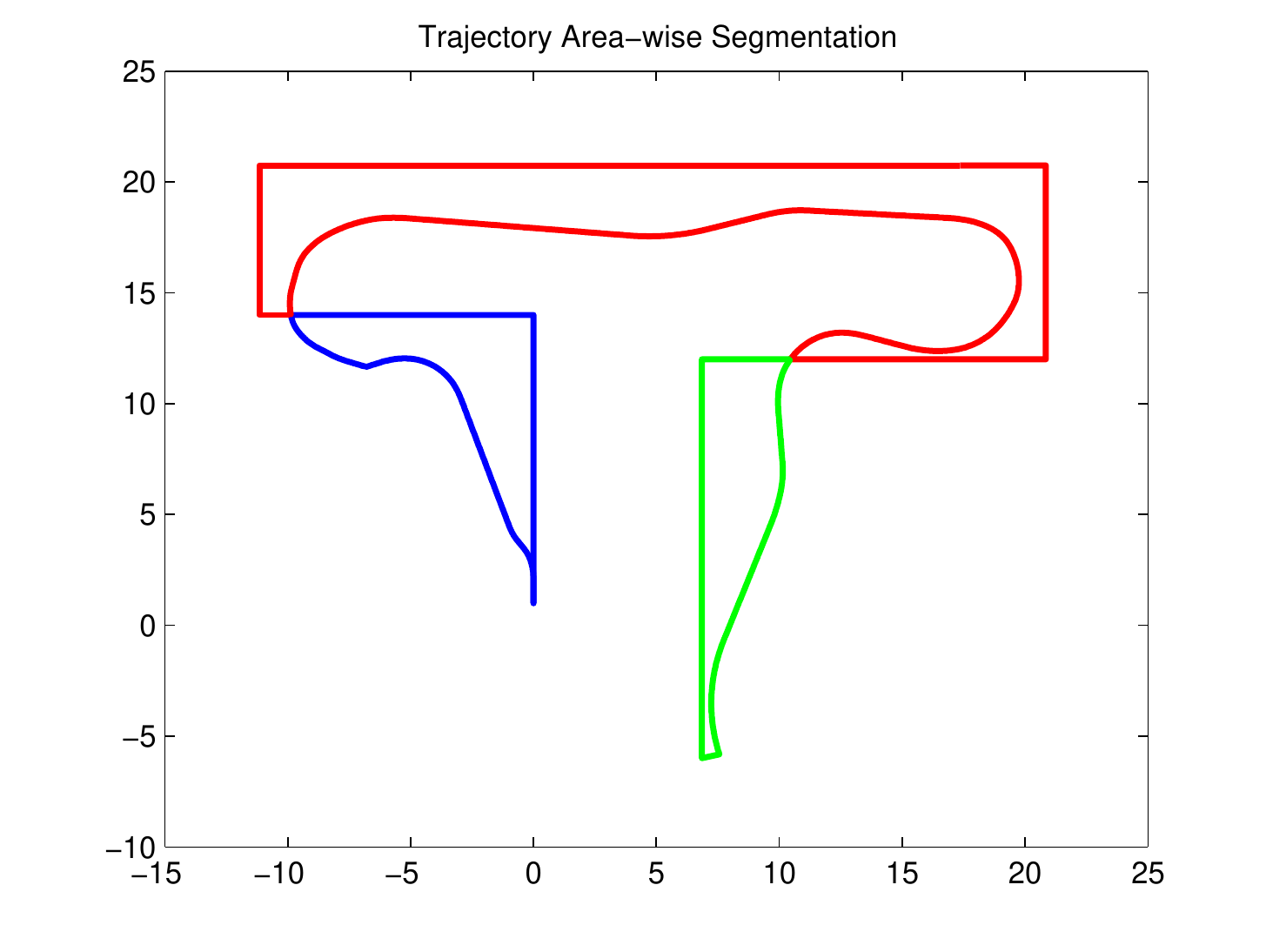}}
\subfigure[]{
\includegraphics[width=0.99\columnwidth]{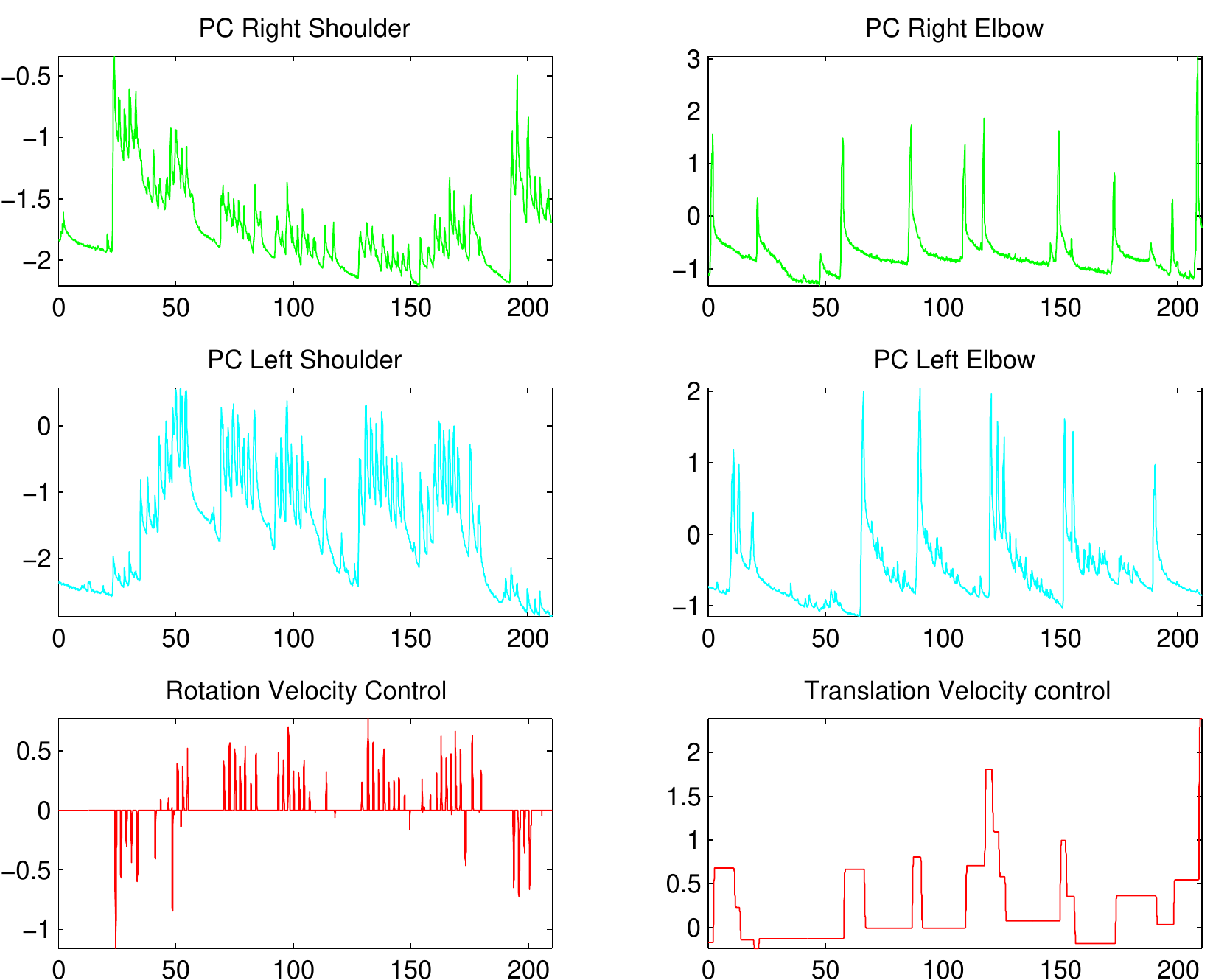}}
\caption{(a) Subject trajectory obtained from one of the experiments. The black circles show the intersection points of both trajectories (b) The subject trajectory is segmented area wise (i.e. in portions, after considering the points of intersections), inorder to calculate the error area. (c) Principal components of shirt's raw signals responsible to produce control signals neccessary for navigating the wheelchair in the virtual reality environment.}
\label{segment}
\end{figure}

\begin{figure*}
\centering
\subfigure[]
{\includegraphics[width=0.8 \columnwidth]{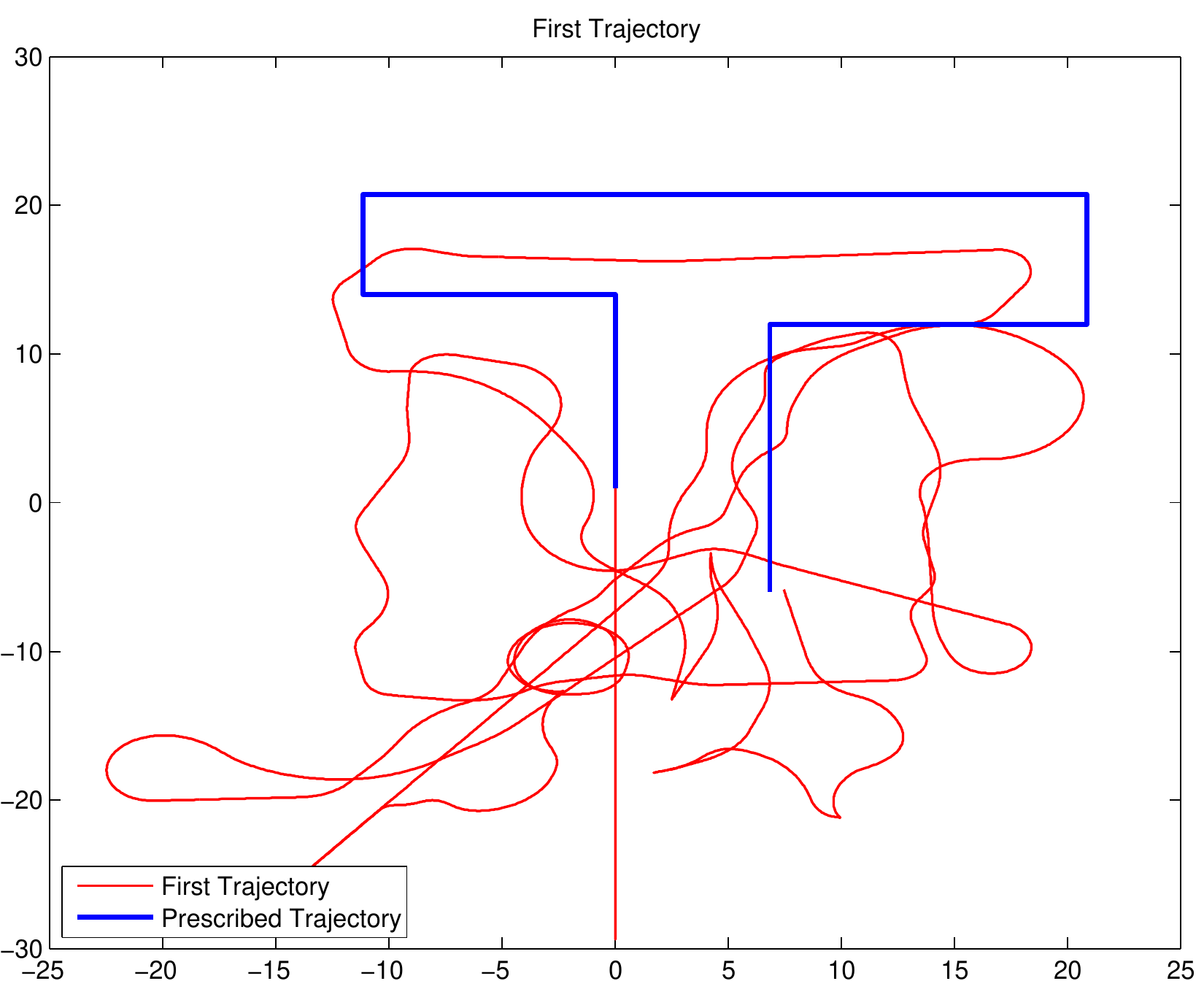}}
\subfigure[]
{\includegraphics[width=0.8 \columnwidth]{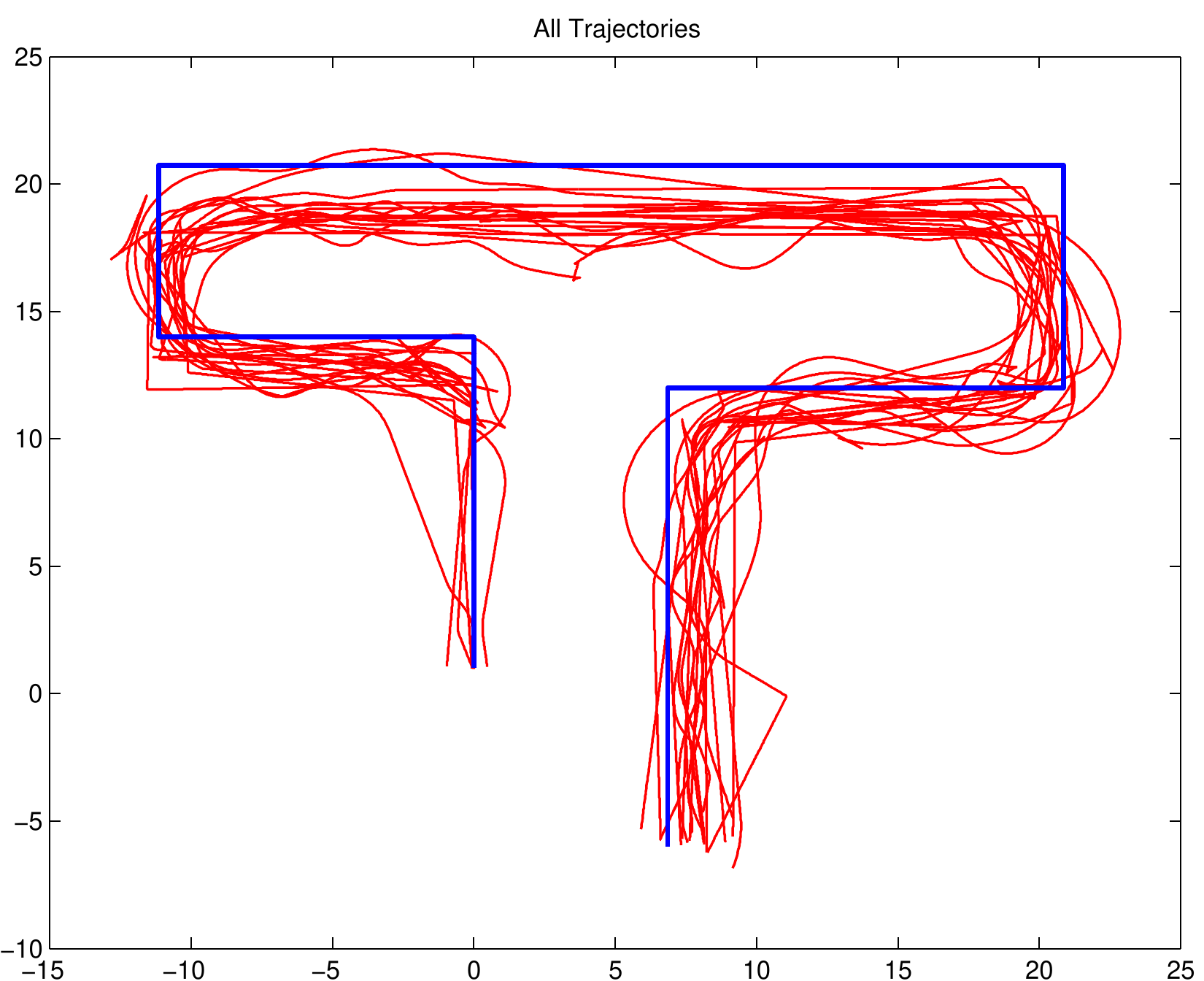}}
\subfigure[]
{\includegraphics[width=0.8 \columnwidth]{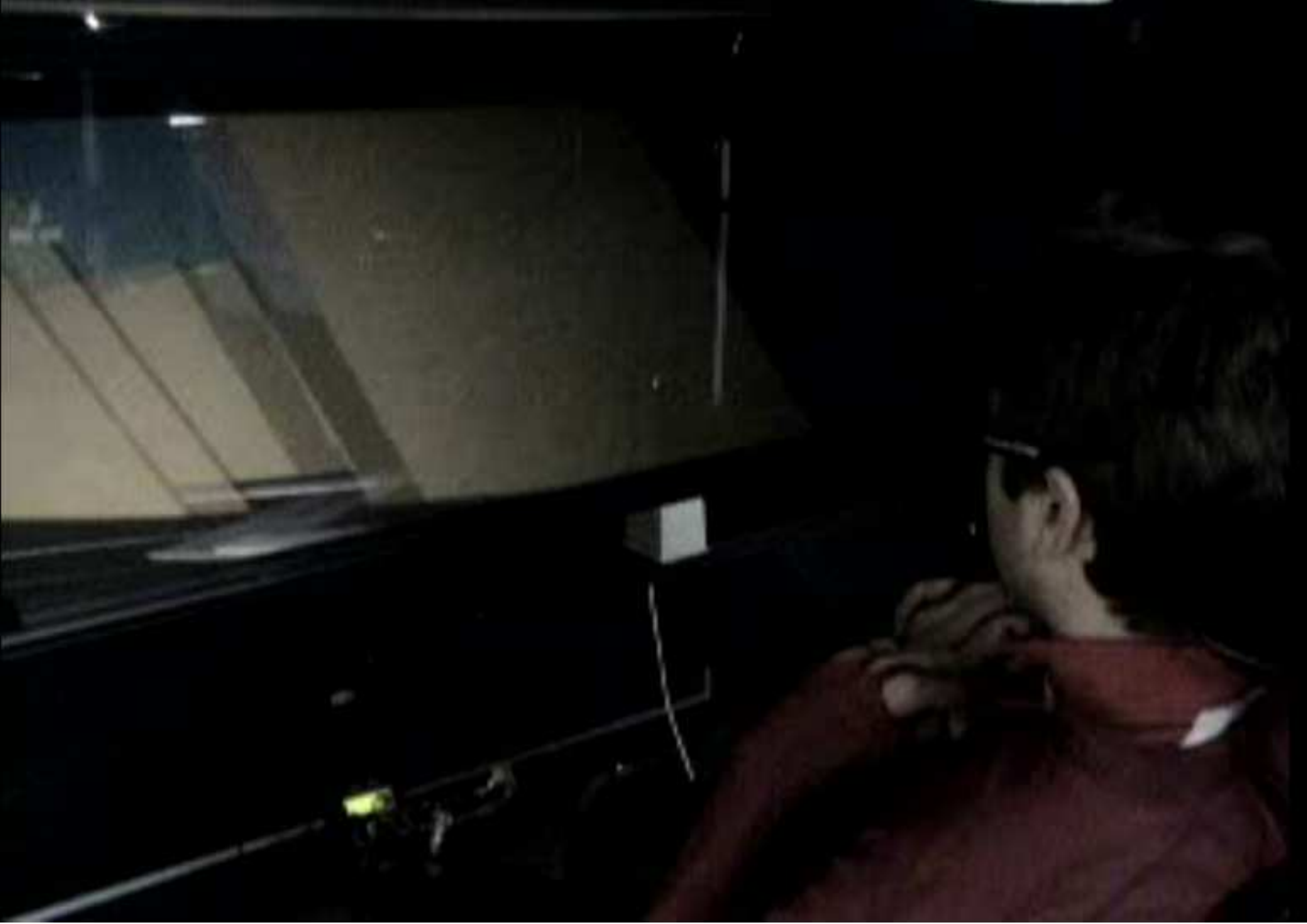}}
\subfigure[]
{\includegraphics[width=0.8 \columnwidth]{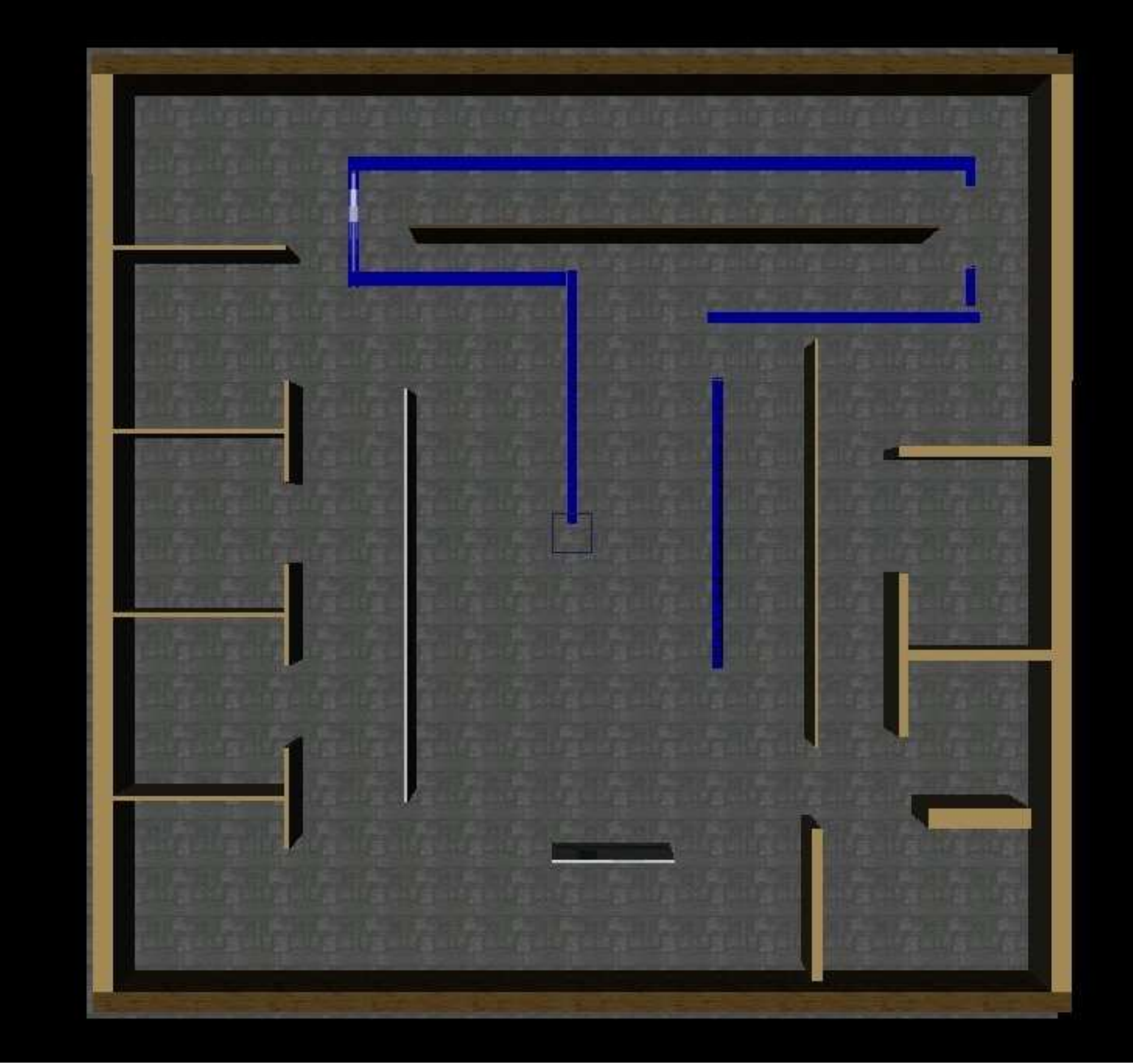}}
\caption{ In (a) red trajectory is obtained on the first day of experiment when subject used the sensor shirt and was asked to move on the marked line (in blue). (b) Shows all trajectories (in red) obtained after 22 preliminary experiment obtained from the subject's travel on the marked line in the immersive virtual environment. In (c) subject sitting infront of Virtual Reality is immersed in the 3D scene and navigating the wheelchair by the residual mobility captured by sensors, along the line marked on the floor. In (d) the top view of the virtual environment is shown with the line (blue coloured) marked on the floor.}
\label{traj}
\end{figure*}

After each trial, the trajectories obtained from the participant's body movements were plotted against the prescribed path. It is important to note here that all of the trajectories were obtained using a uniform sensor shirt control scheme. We analyzed the trajectories using the following measures:
\begin{enumerate}
\item The distance travelled by the subject from the start point to the end point of the prescribed path called $(Dist)$ (shown in fig\ref{segment}(a,b)).
\item The error between the prescribed trajectory and the subject's actual trajectory obtained by calculating the segmented area between both trajectories, from the start point to the end point of the prescribed path called $(E_{diff})$.
\end{enumerate}
In the first trial result shown in fig.\ref{traj}(a) the subject began by familiarizing himself with the control strategy (through arm and shoulder movements) without following the prescribed trajectory. When the subject completed this initial step he started following the prescribed path (also shown in \ref{traj}(a)). The participant moved in different directions in the virtual environment as shown in fig.\ref{traj}(a), to learn the control criteria. As the subject spent more time moving in the virtual scene the understanding of the control map improved and the subject was able to navigate the scene with greater accuracy. 

\begin{figure*}
\centering
\subfigure[]{
\includegraphics[width=0.9\columnwidth]{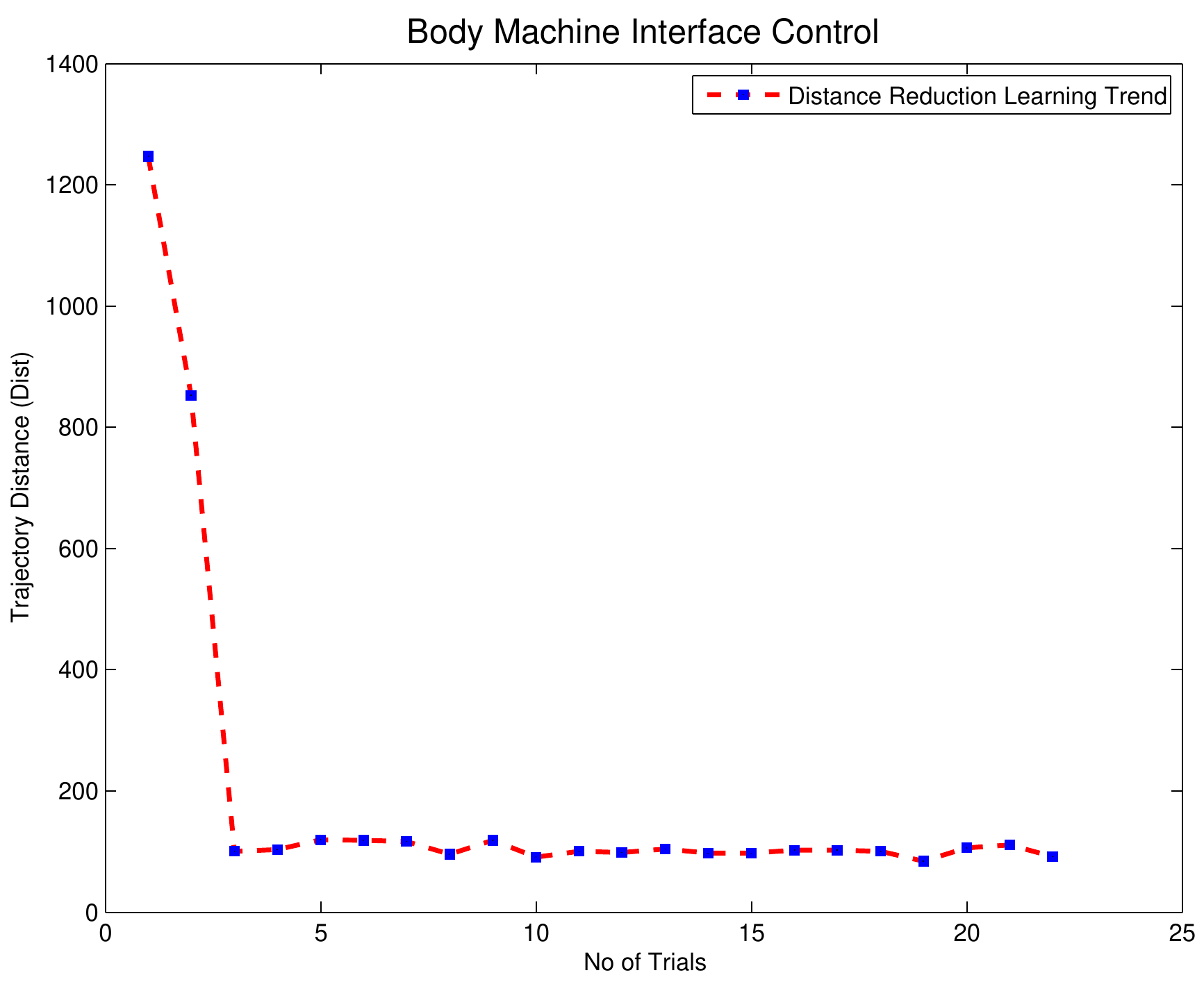}}
\subfigure[]{
\includegraphics[width=0.9\columnwidth]{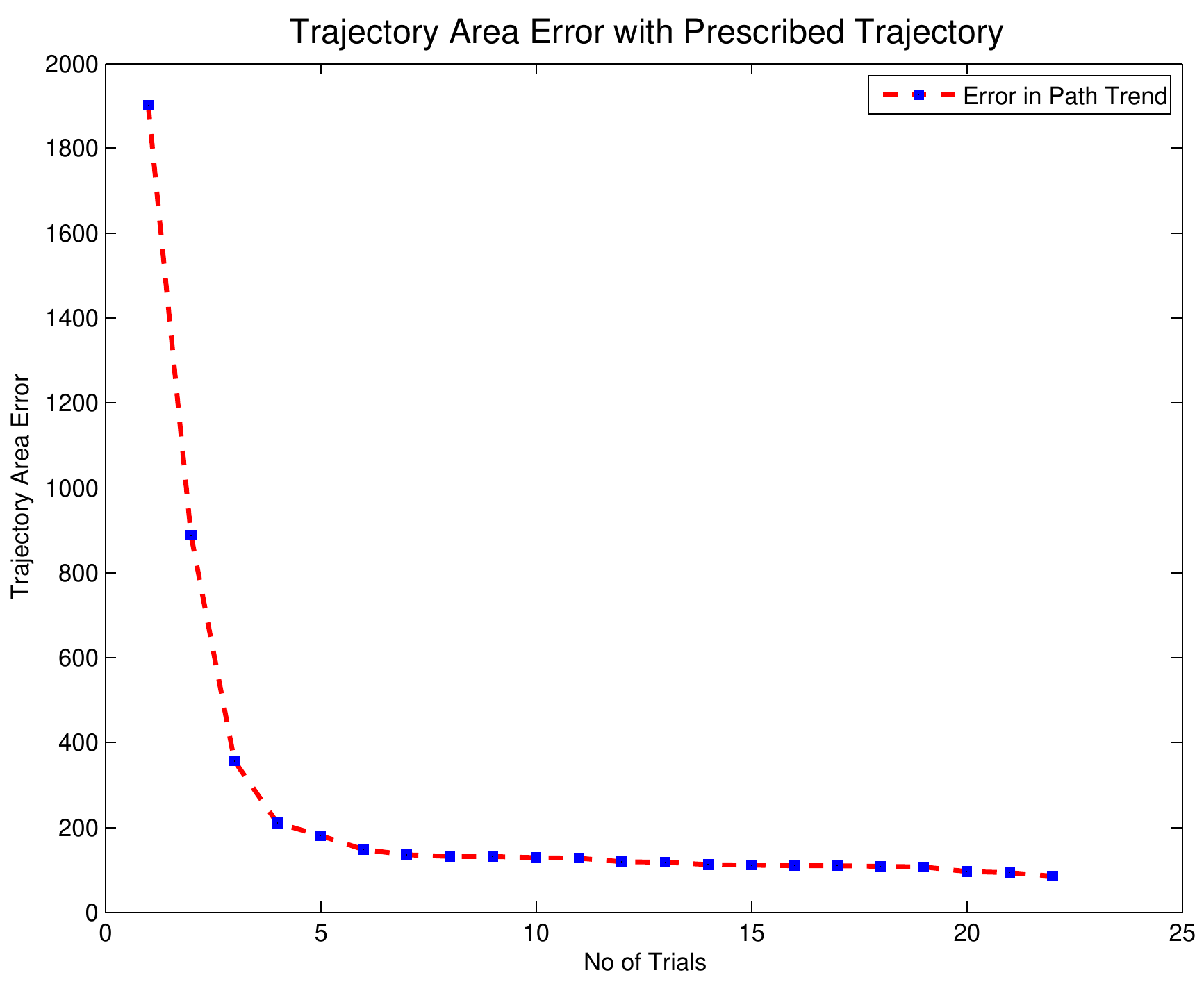}}
\caption{(a) Distance reduction in the trajectories after every trial. (b) Area error measure ($E_{diff}$) between the prescribed and subject's trajectory at each time per day trial.}
\label{error}
\end{figure*}

  The data in fig.\ref{error}(c,d) shows a monotonic reduction in the subject's trajectory error from trial to trial. This is consistent with the hypothesis that, through practice, a subject is able to adapt to their environment using the novel control strategy of moving a wheelchair with shoulder and arm movements. The decreasing error trend is evident for both $E_{diff}$ and $dist$ over all of the trials.  

  The results in fig.\ref{error}(a,b,c\&d) plot the total distance traveled by the subject for each trial to reach the prescribed endpoint from the starting point. The drastic reduction in area and distance error between the first, second and third trials, shows that the subject's initial mobility adjustments are significant. In subsequent trials, the subject's movement adjustments are more finely tuned as the subject's familiarity of the sensor shirt-wheelchair control plan improves resulting in smaller distance errors.
\section{conclusion}
The combination of robotics technology, intelligent interfaces and virtual reality allow us to develop new approaches
to the design of assistive devices. Our approach is based on the key concept that the burden of learning should not fall entirely on the human operator. The field of machine learning has been  rapidly developing in the recent decade and is now sufficiently mature to design interfaces that are capable of learning the user as the user is learning to operate the device. In this case, ``learning the user'' means learning the degrees of freedom that the user is capable to move most efficiently and mapping these degrees of freedom onto wheelchair controls. We should stress that such mapping cannot be static, as in some cases the users will eventually improve with practice. In other, more unfortunate cases, a disability may be progressive and the mobility of the disabled user will gradually deteriorate. In both situations the bodymachine interface must be able to adapt and to update the transformation from body-generated signals to efficient patterns of control.The final aim is to facilitate the formation of new and efficient maps from body motions to operational space.
\section*{Acknowledgment}
This research was supported by NINDS 1R21HD053608, and by a grant of the Craig H. Neilsen Foundation. TG received support from the Macquarie University's Post-graduate research fund.
%
\bibliographystyle{IEEEtran}
\bibliography{./r1,./r2,./r3,./r4,./r5,./r6,./r7,./r8}
\end{document}